\documentclass{article}

\usepackage[square, numbers]{natbib}
\usepackage[final]{neurips_2022}
\usepackage[utf8]{inputenc} 
\usepackage[T1]{fontenc}    
\usepackage{hyperref}       
\usepackage{url}            
\usepackage{booktabs}       
\usepackage{amsfonts}       
\usepackage{nicefrac}       
\usepackage{microtype}      
\usepackage{multirow}       

\usepackage{caption}
\usepackage{subcaption}

\usepackage{wrapfig}

\usepackage{amsmath,amsfonts,bm}

\def\eqref#1{equation~\ref{#1}}

\def\1{\bm{1}}

\def\rmQ{{\mathbf{Q}}}

\DeclareMathAlphabet{\mathsfit}{\encodingdefault}{\sfdefault}{m}{sl}
\SetMathAlphabet{\mathsfit}{bold}{\encodingdefault}{\sfdefault}{bx}{n}

\def\sA{{\mathbb{A}}}

\DeclareMathOperator*{\argmax}{arg\,max}

\usepackage{amssymb} 

\usepackage[ruled,vlined,linesnumbered,noresetcount]{algorithm2e}
\SetKwInput{KwInput}{Input} 
\SetKwInput{KwOutput}{Output} 

\usepackage{enumitem}

\usepackage{bbm}

\usepackage{mathtools}

\usepackage[usenames, dvipsnames, table]{xcolor}

\hypersetup{colorlinks,citecolor=NavyBlue,
linkcolor=NavyBlue,urlcolor=NavyBlue}
\usepackage[nameinlink,capitalise]{cleveref}

\usepackage{adjustbox}

\newcommand{\ourmethod}{Sym-NCO}

\usepackage{lipsum}

\usepackage{amsthm}
\theoremstyle{definition}
\newtheorem{definition}{Definition}[section]
\newtheorem{theorem}{Theorem}[section]

\newcommand{\sol}{{\boldsymbol{\pi}}}
\newcommand{\sym}{\xleftrightarrow{\text{sym}}}

\newcommand{\psl}[0]{\mathcal{L}_{\text{ps}}}
\newcommand{\ssl}[0]{\mathcal{L}_{\text{ss}}} 

\DeclareUnicodeCharacter{0301}{\'{e}}

\title{Sym-NCO: Leveraging Symmetricity for \\Neural Combinatorial Optimization}

\author{Minsu Kim \hspace{10pt} Junyoung Park \hspace{10pt} Jinkyoo Park \\
  Korea Advanced Institute of Science and Technology (KAIST) \\
  Dept. Industrial $\&$ Systems Engineering \\
  \texttt{\{min-su, Junyoungpark, jinkyoo.park\}@kaist.ac.kr}} 

\begin{document}
\maketitle

\begin{abstract}
Deep reinforcement learning (DRL)-based combinatorial optimization (CO) methods (i.e., DRL-NCO) have shown significant merit over the conventional CO solvers as DRL-NCO is capable of learning CO solvers less relying on problem-specific expert domain knowledge (heuristic method) and supervised labeled data (supervised learning method). This paper presents a novel training scheme, \ourmethod{}, which is a regularizer-based training scheme that leverages universal symmetricities in various CO problems and solutions. Leveraging symmetricities such as rotational and reflectional invariance can greatly improve the generalization capability of DRL-NCO because it allows the learned solver to exploit the commonly shared symmetricities in the same CO problem class. Our experimental results verify that our \ourmethod{} greatly improves the performance of DRL-NCO methods in four CO tasks, including the traveling salesman problem (TSP), capacitated vehicle routing problem (CVRP), prize collecting TSP (PCTSP), and orienteering problem (OP), without utilizing problem-specific expert domain knowledge. Remarkably, \ourmethod{} outperformed not only the existing DRL-NCO methods but also a competitive conventional solver, the iterative local search (ILS), in PCTSP at 240$\times$ faster speed. Our source code is available at \url{https://github.com/alstn12088/Sym-NCO}. 
\end{abstract}

\section{Introduction}
Combinatorial optimization problems (COPs), mathematical optimization problems on discrete input space, have been used to solve numerous valuable applications, including vehicle routing problems (VRPs) \cite{cvrp,transportation}, drug discovery \cite{ahn2020guiding,ahn2021spanning}, and semi-conductor design \cite{placement1,placement2,liao2020attention,auto-router,auto-router2}. However, finding an optimal solution to COP is difficult due to its NP-hardness. Therefore, computing near-optimal solutions fast is essential from a practical point of view.

\begin{figure}[t]
    \centering
    \begin{subfigure}[b]{0.45\textwidth}
        \centering
        \includegraphics[width=\textwidth]{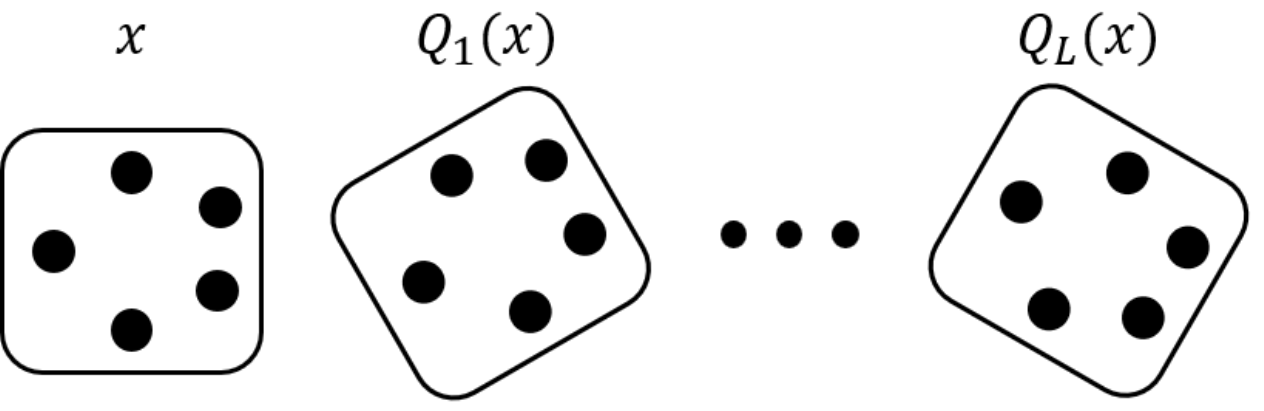}
        \caption{Problem symmetricity}
        \label{figure:problem_sym}
    \end{subfigure}
    \hfill
    \begin{subfigure}[b]{0.45\textwidth}
        \centering
        \includegraphics[width=\textwidth]{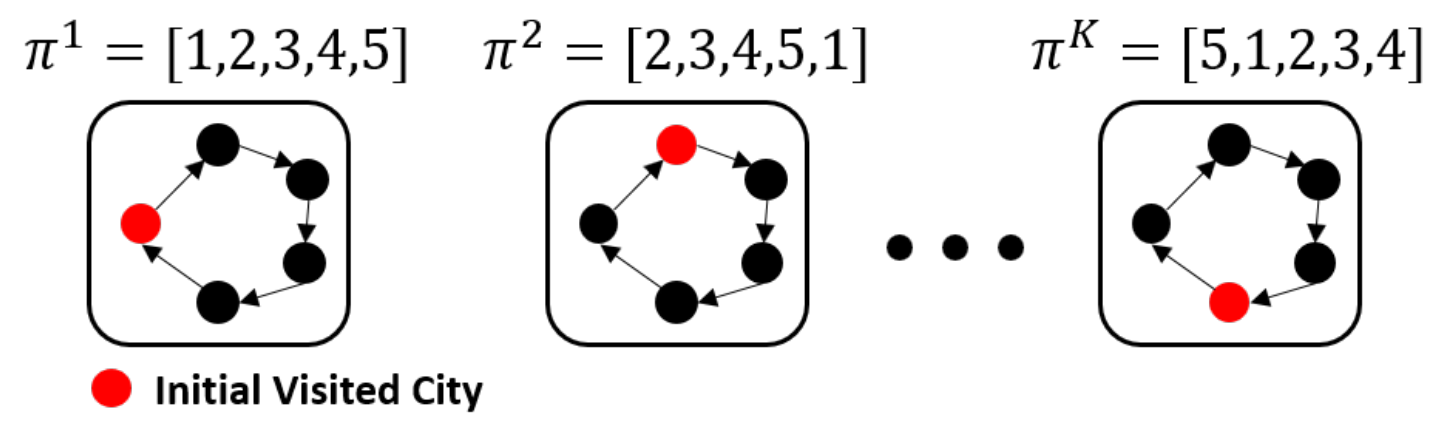}
        \caption{Solution symmetricity}
        \label{figure:sol_sym}
    \end{subfigure}
    \caption{Illustration of symmetricities in CO (exampled in TSP)}     
    \vspace{-2em}
\end{figure}

Conventionally, COPs are solved by integer program (IP) solvers or hand-crafted (meta) heuristics. Recent advances in computing infrastructures and deep learning have conceived the field of neural combinatorial optimization (NCO), a deep learning-based COP solving strategy. Depending on the training scheme, NCO methods are generally classified into supervised learning \cite{pointer,joshi2020learning,kool_dp,fu2020generalize,hottung2020learning} and reinforcement learning (RL) \cite{NLNS,wu2020learning,drl-2opt,mis,kim2021learning,bello2017neural,kool2018attention,Nazari,kwon2020pomo,park2021learning, ma2021learning,xin2021multi, park2021schedulenet}. Depending on the solution generation scheme, NCO methods are also classified into improvement \cite{drl-2opt, wu2020learning, NLNS, chen2019learning, mis, kim2021learning, ma2021learning} and constructive heuristics \cite{bello2017neural,kool2018attention,Nazari,kwon2020pomo,park2021learning,xin2021multi, park2021schedulenet}. Among the NCO approaches, deep RL (DRL)-based constructive heuristics (i.e., DRL-NCO) are favored over conventional approaches for two major reasons. First, RL can be applied to train the NCO model in less explorered CO tasks because training RL does not require domain expert knowledge and supervised labels from a verified solver. Second, it is easy to produce qualified feasible solutions because the constructive process can easily avoid constraint-violated actions \cite{kool2018attention}. Despite the strength of DRL-NCO, there exists a performance gap between the state-of-the-art conventional heuristics and DRL-NCO. In an effort to close the gap, there have been attempts to employ problem-specific heuristics to existing DRL-NCO methods \cite{kwon2020pomo,wang2021rewriting}. However, devising a general training scheme to improve the performance of DRL-NCO still remains challenging. 
\begin{wrapfigure}{r}{0.5\textwidth}
  \begin{center}
    \includegraphics[width=0.5\textwidth]{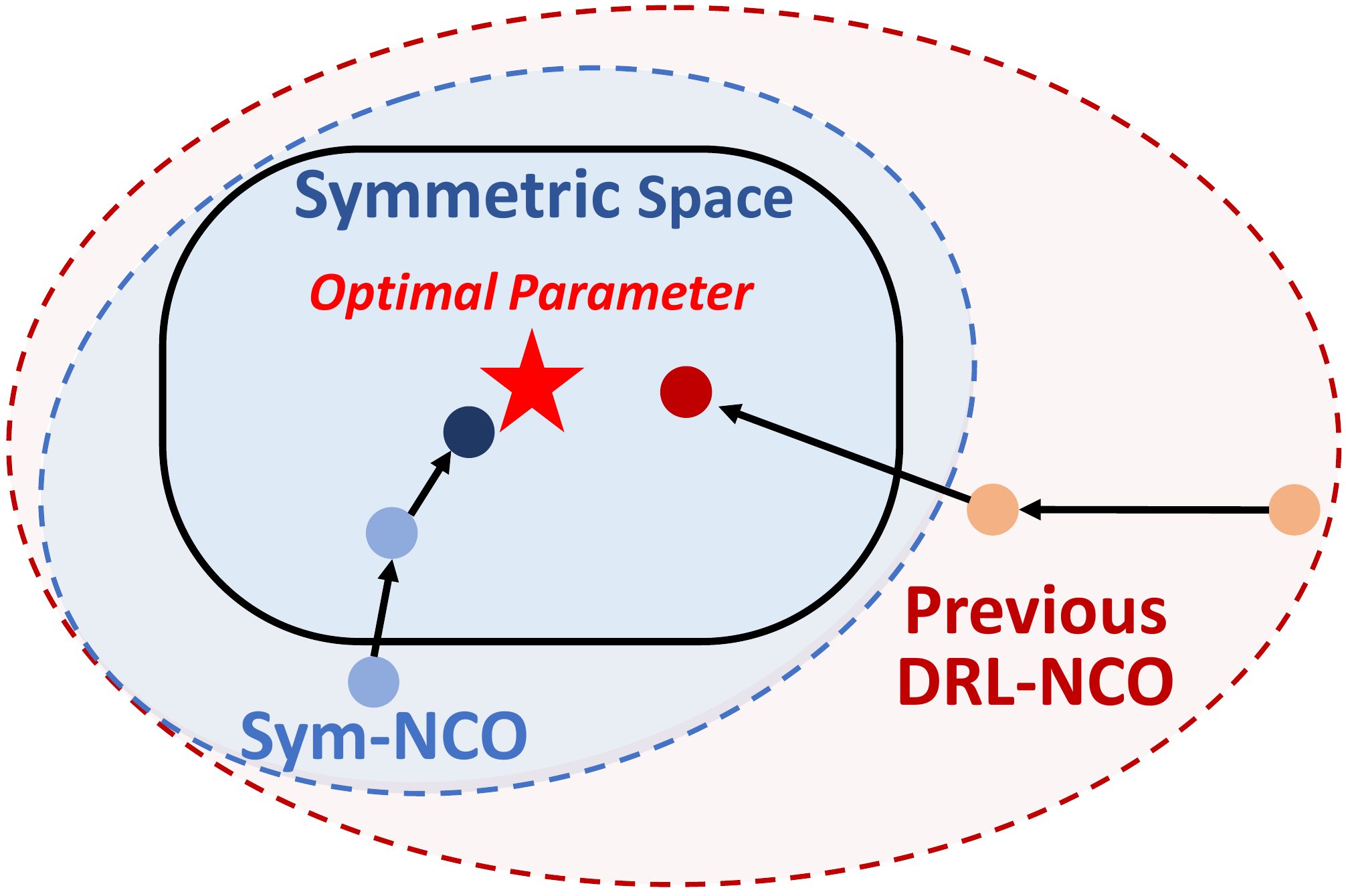}
  \end{center}
  \caption{Illustration that describes an advantage of Sym-NCO. An optimal training parameter is in symmetric space. Sym-NCO makes a more compact training space compared with previous DRL-NCO and supports the NCO model efficiently converges in near-optimal parameters.  
   }
  \label{figure:motivation}
  \vspace{-1.5em}
\end{wrapfigure}
 In this study, we propose the Symmetric Neural Combinatorial Optimization (\ourmethod{}), a general training scheme applicable to universal CO problems. \ourmethod{} is a regularization-based training scheme that leverages the symmetricities commonly found in COPs to increase the performance of existing DRL-NCO methods.  \ourmethod{} leverages two types of symmetricities innate in COP that are defined on the Euclidean graph. First, the problem symmetricity is derived from the rotational invariance of the solution; the rotated graph must exhibit the same optimal solution as the original graph as shown in \cref{figure:problem_sym}. Second, the solution symmetricity refers to the property that solutions have identical output values (See \cref{figure:sol_sym}). 
 

To train an effective NCO solver while leveraging the symmetricities, we employ REINFORCE algorithm with the baselines terms specially designed to impart solution and problem symmetricities. REINFORCE algorithm is used because the well-known effective NCO solvers are trained by REINFORCE; thus, we can improve such solvers by just modifying their baseline with our symmetricity-considered baseline terms. Specifically, we sample multiple solutions from the transformed problems and use the average return of them. Then, REINFORCE pushes each solution sampled by the solver to excel this baseline during training, thus improving the policy and making all the solutions the same, i.e., the problem and solution symmetricities are realized.


\textbf{Motivation for learning symmetricity.} Leveraging symmetricity is important to train CO models for two major reasons. Firstly, symmetricity is a strong inductive bias that can support the training process of DRL by making compact training space as shown in \cref{figure:motivation}. Secondly,
learning symmetricity is beneficial to increasing generalization capability for unseen CO problems because symmetricity induces the invariant representation that every COP contains.

\textbf{Novelty.} The major novelty of the proposed learning strategy is that it can easily improve existing powerful CO models; existing equivariant neural network schemes must be re-designed at the architecture level for adapting to CO.

\section{Symmetricity in Combinatorial Optimization Markov Decision Process}
This section presents several symmetric characteristics found in combinatorial optimization, which is formulated in the Markov decision process. The objective of NCO is to train the $\theta$-parameterized solver $F_\theta$ by solving the following problem:
\begin{align}
    {\theta}^*= \argmax_\theta{\mathbb{E}}_{\boldsymbol{P}\sim \rho}\big[{\mathbb{E}}_{\textcolor{black}{\sol \sim F_\theta(\boldsymbol{P})}}\big[R(\sol;\boldsymbol{P})\big] \big]
    \label{eqn:objective}
\end{align}
where $\boldsymbol{P}=(\boldsymbol{x},\boldsymbol{f})$ is a problem instance with the $N$ node coordinates $\boldsymbol{x}=\{x_i\}_{i=1}^N$ and corresponding $N$ features $\boldsymbol{f} = \{f_i\}_{i=1}^N$. The $\rho$ is a problem generating distribution. The $\sol = \{\pi_i\}_{i=1}^N$ is a solution where each element is index value $\pi_i \in \{1,...,N\}$ for coordinates $\boldsymbol{x}=\{x_i\}_{i=1}^N$. The $R(\sol;\boldsymbol{P})$ is objective value for $\sol$ on problem $\boldsymbol{P}$. 

For example, assume that we solve TSP with five cities (i.e. $N=5$). Then the problem instance $\boldsymbol{P}$ contains five city coordinates $\boldsymbol{x}=\{x_i\}_{i=1}^5$ where the salesman must visit. The solution $\sol$ is a sequence of city indices; if $\sol = \{1,3,2,5,4\}$, the salesman visits city coordinates as $x_1 \rightarrow x_3 \rightarrow x_2 \rightarrow x_5 \rightarrow x_4 \rightarrow x_1$ (the salesman must go back to first visited city to complete a tour). In TSP, each city contains the homogeneous features $\boldsymbol{f}$. Thus, objective $R$ of the TSP is defined as the negative of tour length: $R(\sol;\boldsymbol{P}) = -\big( \sum_{i=1}^{4}{||x_{\pi_{i+1}}-x_{\pi_{i}}||} + ||x_{\pi_{5}}-x_{\pi_{1}}|| \big)$. This combinatorial decision process can be expressed as Markov decision process, and the solver $F_{\theta}(\sol|\boldsymbol{P})$ can be expressed as \textit{instance conditioned policy}. To this end, we can utilize deep reinforcement learning for training solver $F_{\theta}(\sol|\boldsymbol{P})$. We formally define the Markov decision process for CO in the below chapter.



\subsection{Combinatorial optimization Markov decision process}
We define the combinatorial optimization Markov decision process (CO-MDP) as the sequential construction of a solution of COP. For a given $\boldsymbol{P}$, the components of the corresponding CO-MDP are defined as follows:
\vspace{-0.25cm}
\begin{itemize}[leftmargin=0.5cm]
    \item \textbf{State.} The state $\boldsymbol{s}_t=(\boldsymbol{a}_{1:t},\boldsymbol{x},\boldsymbol{f})$ is the $t$-th (partially complete) solution, where $\boldsymbol{a}_{1:t}$ represents the previously selected nodes.  The initial and terminal states $\boldsymbol{s}_0$ and $\boldsymbol{s}_T$ are equivalent to the empty and completed solution, respectively. In this paper, we denote the solution $\sol(\boldsymbol{P})$ as the completed solution.
    \item \textbf{Action.} The action $a_t$ is the selection of a node from the un-visited nodes (i.e., $a_t \in \sA_t =\{ \{1,...,N\} \setminus \{\boldsymbol{a}_{1:t-1}\}\}$). 
    \item \textbf{Reward.} The reward function $R(\sol;\boldsymbol{P})$ maps the objective value from given $\sol$ of problem $\boldsymbol{P}$. We assume that the reward is a function of $\boldsymbol{a}_{1:T}$ (solution sequences), $||x_i - x_j||_{i,j \in \{1,...N\}}$ (relative distances) and $\boldsymbol{f}$ (nodes features). In TSP, capacitated VRP (CVRP), and prize collecting TSPs (PCTSP), the reward is the negative of the tour length. In orienteering problem (OP), the reward is the sum of the prizes.
\end{itemize}
Having defined CO-MDP, we define the solution mapping as follows: $\sol \sim F_{\theta}(\cdot|P) = \prod_{t=1}^{T} p_{\theta}(a_t|\boldsymbol{s}_t)$ where $p_\theta(a_t|\boldsymbol{s}_t)$ is the policy that produces $a_t$ at $\boldsymbol{s}_t$, and $T$ is the maximum number of states in the solution construction process.

\subsection{Symmetricities in CO-MDP}
\label{subsec:sym-co-mdp}

Symmetricities are found in various COPs. We conjecture that imposing those symmetricities on $F_\theta$ improves the generalization and sample efficiency of $F_\theta$. We define the two identified symmetricities that are commonly found in various COPs: 

\begin{definition}[\textbf{Problem Symmetricity}] Problem $\boldsymbol{P}^i$ and $\boldsymbol{P}^j$ are problem symmetric ($\boldsymbol{P}^i \sym \boldsymbol{P}^j$) if their optimal solution sets are identical.
\label{def:pro_sym}
\end{definition}


\begin{definition}[\textbf{Solution Symmetricity}] Two solutions $\sol^i$ and $\sol^j$ are solution symmetric ($\sol^i \sym \sol^j$) on problem $\boldsymbol{P}$ if
$R(\sol^i;\boldsymbol{P}) = R(\sol^j;\boldsymbol{P})$.
\label{def:sol_sym}
\end{definition}



An exemplary problem symmetricity found in various COPs is the rotational symmetricity:
\begin{theorem}[\textbf{Rotational symmetricity}] For any orthogoanl matrix $Q$, the problem $\boldsymbol{P}$ and $Q(\boldsymbol{P}) \triangleq \{\{Qx_i\}_{i=1}^{N},\boldsymbol{f}\}$ are problem symmetric: i.e., $\boldsymbol{P} \sym Q(\boldsymbol{P})$. See \cref{append: proof} for the proof.
\label{thm:rot_sym}
\end{theorem}

Rotational problem symmetricity is identified in every Euclidean COPs. On the other hand, solution symmetricity cannot be identified easily as the properties of the solutions are distinct for every COP. 






\section{Symmetric Neural Combinatorial Optimization}
\label{sec:sym-nco}
\begin{figure}[t]
\centering
\includegraphics[width=0.8\textwidth]{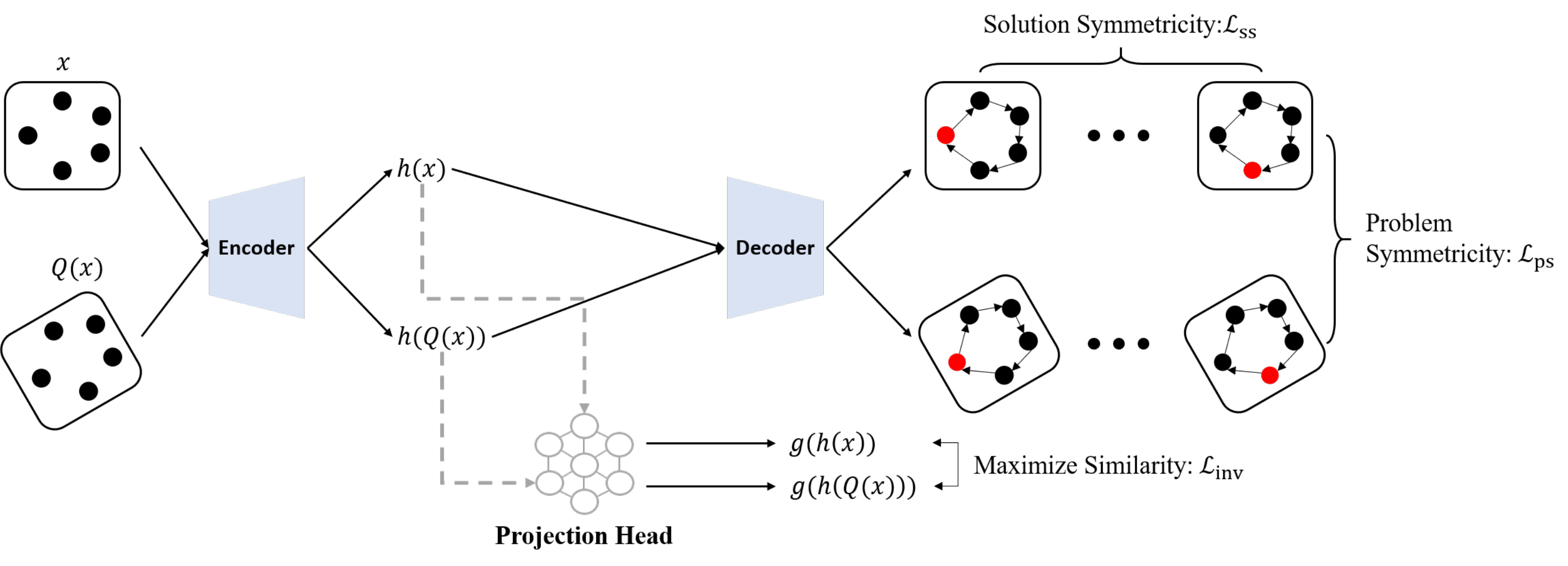}
\caption{An overview of \ourmethod{}}
\vspace{-2em}
\end{figure}

This section presents \ourmethod{}, an effective training scheme that leverages the symmetricities of COPs. \ourmethod{} learns a solve $F_\theta$ by minimizing the total loss function:
\begin{align}
    \mathcal{L_{\text{total}}}= 
    \textcolor{NavyBlue}{\mathcal{L}_{\text{Sym-RL}}} + \alpha \mathcal{L}_{\text{inv}} = \textcolor{NavyBlue}{\psl + \beta \ssl} + \alpha \mathcal{L}_{\text{inv}}    
\end{align}
where $\mathcal{L}_{\text{Sym-RL}}$ is REINFORCE loss term supported by symmetricity and $\mathcal{L}_{\text{inv}}$ is the regularization loss to induce that invariant representation. The $\mathcal{{L}_{\text{Sym-RL}}}$ is composed with $\ssl$, the REINFORCE loss term of \cref{eqn:objective} with solution symmetricity regularizing mechanism, and $\psl$, the REINFORCE loss term of \cref{eqn:objective} with both solution and problem symmetricity regularization. $\alpha, \beta \in [0,1]$ are the weight coefficients. In the following subsections, we explain each loss term in detail.

\subsection{Regularizing REINFORCE with problem and solution symmetricities via $\mathcal{{L}_{\text{Sym-RL}}}$}

As discussed in \cref{subsec:sym-co-mdp}, COPs have problem and solution symmetricities. We explain how to learn solver $F_{\theta}$ using REINFORCE with the specially designed baseline to approximately impose symmetricities. We provide the policy gradients to $\ssl\left(\sol(\boldsymbol{P})\right)$ and $\psl\left(\sol(\boldsymbol{P})\right)$ in the context of the REINFORCE algorithm \cite{williams1992simple} with the proposed baseline schemes. 

\textbf{Leveraging solution symmetricity.} As defined in \cref{def:sol_sym}, the symmetric solutions must have the same objective values. We propose the REINFORCE loss $\ssl$ with the baseline specially designed to exploit the solution symmetricity of CO as follows:
\begin{align}
    \ssl &= -\mathbb{E}_{\sol \sim F_\theta(\cdot|\boldsymbol{P})}
            \big[R(\sol;\boldsymbol{P})
            \big] \\
    \nabla_\theta  \ssl &= -
        \mathbb{E}_{\sol \sim F_\theta(\cdot|\boldsymbol{P})}
        \Big[
            \big[R(\sol;\boldsymbol{P})-b(\boldsymbol{P})
            \big]
            \nabla_\theta \log F_\theta \Big] \\ &\approx -
       \frac{1}{K}\sum_{k=1}^{K}
        \Big[
            \big[R(\textcolor{NavyBlue}{\sol^{k}};\boldsymbol{P})-\frac{1}{K}\sum_{k=1}^{K}R
                    (\textcolor{NavyBlue}{\sol^{k}};\boldsymbol{P})
            \big]
            \nabla_\theta \log F_\theta \Big]
\end{align}
where $\{\textcolor{NavyBlue}{\sol^{k}}\}_{k=1}^{K}$ are the solutions of $P$ sampled from $F_{\theta}(\sol|\boldsymbol{P})$, $\log F_\theta$ is the log-likelihood of $F_\theta$, $K$ is the number of sampled solutions, $b(\boldsymbol{P})$ is a shared baseline which is the average reward from $K$ solutions for the identical problem $\boldsymbol{P}$.  

The $\ssl$ approximately imposes solution symmetricity using REINFORCE algorithm with a novel baseline $b(\boldsymbol{P})$. The sum of advantage in the solution group $\{\textcolor{NavyBlue}{\sol^{k}}\}_{k=1}^{K}$ is always zero:
\begin{align}
\frac{1}{K}\sum_{k=1}^{K}
        \Big[
            \big[R(\textcolor{NavyBlue}{\sol^{k}};\boldsymbol{P})-\frac{1}{K}\sum_{k=1}^{K}R
                    (\textcolor{NavyBlue}{\sol^{k}};\boldsymbol{P})
            \big]\Big] = 0     
\end{align}
The $\ssl$ induces competition among the rewards, $R(\sol^{1};\boldsymbol{P}),...,R(\sol^{K};\boldsymbol{P})$ which can be seen as a zero-sum game. Therefore, $\ssl$ improves the overall reward quality of the solution group using the proposed competitive REINFORCE scheme, making the solver generate high-rewarded solutions but a small reward deviation between solutions. The small reward-deviation indicates $\ssl$ approximately imposes solution symmetricity to solver $F_{\theta}$.



The POMO \cite{kwon2020pomo} employed a similar training technique with our $\ssl$, that finds symmetric solutions by forcing $F_\theta$ to visit all possible initial cities when solving TSP and CVRP. However, the reward of COPs, including CVRP, PCTSP, and OP, is usually sensitive to first-city selection. Therefore, POMO can be an excellent standalone method for TSP but can be further improved using our $\ssl$ loss term in other tasks including CVRP.




\textbf{Leveraging problem symmetricity.} As discussed in \cref{subsec:sym-co-mdp}, the rotational problem symmetricity is common in various COPs. We propose the REINFORCE loss $\psl$ which is equipped with problem symmetricity: \begin{align}
    \psl &= -\mathbb{E}_{Q^l \sim \rmQ}
        \mathbb{E}_{\sol \sim F_\theta(\cdot|Q^l(\boldsymbol{P}))}
            \big[R(\sol;\boldsymbol{P})
            \big]\\
    \nabla_\theta \psl &= -\mathbb{E}_{Q^l \sim \rmQ}
    \bigg[
        \mathbb{E}_{\sol \sim F_\theta(\cdot|Q^l(\boldsymbol{P}))}
        \Big[
            \big[R(\sol;\boldsymbol{P})-b(\boldsymbol{P},\boldsymbol{Q})  
            \big]
            \nabla_\theta \log F_\theta \Big] \bigg]
            \\ &\approx \frac{1}{LK}\sum_{l=1}^{L}\sum_{k=1}^{K} \Big[
            \big[R(\textcolor{NavyBlue}{\sol^{l,k}};\boldsymbol{P})-\frac{1}{LK}\sum_{l=1}^{L}\sum_{k=1}^{K}R
                    (\textcolor{NavyBlue}{\sol^{l,k}};\boldsymbol{P})  
            \big]
            \nabla_\theta \log F_\theta \Big] \bigg]  
\end{align}

where $\rmQ$ is the distribution of random orthogonal matrices, $Q^l$ is the $l^{\text{th}}$ sampled rotation matrix, and $\textcolor{NavyBlue}{\sol^{l,k}}$ is the $k^{\text{th}}$ sample solution of the $l^{\text{th}}$ rotated problem. We construct $L$ problem symmetric problems, $Q^{1}(\boldsymbol{P}),...,Q^{L}(\boldsymbol{P})$, by using the sampled rotation matrices, and smaple $K$ symmetric solutions from each of the $L$ problems. Then, the shared baseline $b(\boldsymbol{P},\boldsymbol{Q})$ is constructed by averaging $L \times K$ solutions.

Similar to the regularization scheme of $\ssl$, the advantage term of $\psl$ also induces competition between solutions sampled from rotationally symmetric problems. Since the rotational symmetricity is defined such that $x$ and $Q_l(x)$ have the same solution, the negative advantage value forces the solver to find a better solution. As mentioned in \cref{subsec:sym-co-mdp}, problem symmetricity in COPs is usually pre-identified (i.e. there is provable guaranteed symmetricity such as rotational symmetricity \cref{thm:rot_sym}); $\psl$ are applicable to general COPs. Moreover, multiple solutions are sampled for each symmetric problem so that $\psl$ can also identify and exploit the solution symmetricity with a similar approach taken for $\ssl$. We provide detailed implementation and design guides regarding the integration strategy of $\ssl$ and $\psl$ in \cref{append:deisgn_guide}.




\subsection{Learning invariant representation with Pre-identified Symmetricity: $\mathcal{L}_{\text{inv}}$.}\label{subsec:invariant}
By \cref{thm:rot_sym}, the original problem $x$ and its rotated problem $Q(x)$ have identical solutions. Therefore the encoder of $F_\theta$ can be enforced to have invariant representation by leveraging the pre-identified symmetricity: rotation symmetricity. 

We denote $h(x)$ and $h\left(Q(x)\right)$ as the hidden representations of $x$ and $Q(x)$, respectively. To impose the rotational invariant property on $h(x)$, we train solver $F_{\theta}$ with an additional regularization loss term $\mathcal{L}_{\text{inv}}$ defined as:
\begin{align}
    \mathcal{L}_{\text{inv}} = -S_\text{cos}\bigg(g\Big(h(x)\Big), g\Big(h\big(Q(x)\big)\Big)\bigg)
    \vspace{-4pt}
\end{align}
where $S_\text{cos}(a,b)$ is the cosine similarity between $a$ and $b$. $g$ is the MLP-parameterized projection head.

For learning the invariant representation on rotational symmetricity, we penalize the difference between the projected representation $g(h(x))$ and $g(h(Q(x)))$, instead of directly penalizing the difference between $h(x)$ and $h(Q(x)$. This penalizing scheme allows the use of an arbitrary encoder network architecture while maintaining the diversity of $h$ \cite{chen2020simple}. We empirically verified that this approach attains stronger solvers as described in \cref{subsec:hard_const}.



\section{Related Works}

\paragraph{Deep construction heuristics.} Bello et al. \cite{bello2017neural} propose one of the earliest DRL-NCO methods, based on pointer network (PointerNet) \cite{pointer}, and trained it with an actor-critic method. Attention model (AM) \cite{kool2018attention} successfully extends \cite{bello2017neural} by swapping PointerNet with Transformer \cite{transformer}, and it is currently the \textit{de-facto} standard method for NCO. Notably, AM verifies its problem agnosticism by solving several classical routing problems and their practical extensions \cite{liao2020attention,kim2021learning}. The multi-decoder AM (MDAM) \cite{xin2021multi} extends AM by employing an ensemble of decoders. However, such an extension is inapplicable for stochastic routing problems. The policy optimization for multiple optimal (POMO) \cite{kwon2020pomo} extends AM by exploiting the solution symmetricities in TSP and CVRP. Even though POMO shows significant improvements from AM, it relies on problem-specific solution symmetricities for TSP. Our method can be seen as a general-purpose symmetric learning scheme, extended from POMO, which can be applied in more general CO tasks.  

\paragraph{Equivariant deep learning.} In deep learning, symmetricities are often enforced by employing specific network architectures. Niu et al. \cite{niu2020permutation} proposes a permutation equivariant graph neural network (GNN) that produces equivariant outputs to the input order permutations. The $SE(3)$-Transformer \cite{fuchs2020se} restricts the Transformer so that it is equivariant to $SE(3)$ group input transformation. Similarly, equivariant GNN (EGNN) \cite{satorras2021n} proposes a GNN architecture that produces $O(n)$ group equivariant output. These network architectures can dramatically reduce the search space of the model parameters. Some research applies equivariant neural networks to RL tasks to improve sample efficiency \cite{van2020mdp}. Also, there are several works that exploited the symmetric nature of CO. Ouyang et al. \cite{ouyang2021generalization} proposed equivariant encoding scheme by giving rule-based input transformation to input graph. Hudson et al. \cite{hudson2021graph} suggested a line-graph embedding scheme, which is beneficial to process CO graphs with rotational equivariant. Our Sym-NCO is a regularization method that is capable of learning existing powerful CO models without giving rule-based hard constraints to the structure. We empirically study the benefit of Sym-NCO over other symmetricity-based approaches in \cref{subsec:hard_const} and \cref{sym_literature}.

\section{Experiments}
This section provides the experimental results of \ourmethod{} for TSP, CVRP, PCTSP, and OP. Focusing on the fact that \ourmethod{} can be applied to any encoder-decoder-based NCO method, we implement \ourmethod{} on top of POMO \cite{kwon2020pomo} to solve TSP and CVRP, and AM \cite{kool2018attention} to solve PCTSP and OP, respectively. We additionally validate the effectiveness of \ourmethod{} on PointerNet \cite{pointer} at TSP ($N=100$). 

\subsection{Tasks and baseline selections}
TSP aims to find the Hamiltonian cycle with a minimum tour length. We employ Concorde \cite{concorde} and LKH-3 \cite{lkh2017} as the non-learnable baselines, and PointerNet \cite{pointer}, the structured-to-vector deep-Q-network (S2V-DQN) \cite{Khalil}, AM \cite{kool2018attention}, POMO \cite{kwon2020pomo} and MDAM \cite{xin2021multi} as the neural constructive baselines.

CVRP is an extension of TSP that aims to find a set of tours with minimal total tour lengths while satisfying the capacity limits of the vehicles. We employ LKH-3 \cite{lkh2017} as the non-learnable baselines, and Nazari et al. \cite{Nazari}, AM \cite{kool2018attention}, POMO \cite{kwon2020pomo} and MDAM \cite{xin2021multi} as the constructive neural baselines.

PCTSP is a variant of TSP that aims to find a tour with minimal tour length while satisfying the prize constraints. 
We employ the iterative local search (ILS) \cite{kool2018attention} as the non-learnable baseline, and AM \cite{kool2018attention} and MDAM \cite{xin2021multi} as the constructive neural baselines.

OP is a variant of TSP that aims to find the tour with maximal total prizes while satisfying the tour length constraint. We employ \textit{compass} \cite{kobeaga2018efficient} as the non-learnable baseline, and AM \cite{kool2018attention} and MDAM \cite{xin2021multi} as the constructive neural baselines.

\subsection{Experimental setting}
\label{sec:exp_setting}
\textbf{Problem size.} We provide the results of problems with $N=100$ for the four problem classes, and real-world TSP problems with $50<N<250$ from TSPLIB. 

\paragraph{Hyperparameters.} We apply \ourmethod{} to POMO, AM, and PointerNet. To make fair comparisons, we use the same network architectures and training-related hyperparameters from their original papers to train their \ourmethod{}-augmented models. Please refer to Appendix \cref{append: hyperparameter} for more details.

\paragraph{Dataset and Computing Resources.} We use the benchmark dataset \citep{kool2018attention} to evaluate the performance of the solvers. To train the neural solvers, we use \textit{Nvidia} A100 GPU. To evaluate the inference speed, we use an \textit{Intel} Xeon E5-2630 CPU and \textit{Nvidia} RTX2080Ti GPU to make fair comparisons with the existing methods as proposed in \cite{xin2021multi}.

\subsection{Performance metrics}
This section provides detailed performance metrics:

\textbf{Average cost.} We report an average cost of 10,000 benchmark instances which is proposed by \cite{kool2018attention}.

\textbf{Evaluation speed.} We report the evaluation speeds of solvers in a out-of-the-box manner as they are used in practice. In that regard, the execution time of non-neural and neural methods are measured on CPU and GPU, respectively.

\textbf{Greedy/Multi-start performance.} For neural solvers, it is a common practice to measure \textit{multi-start} performance as its final performance. However, when those are employed in practice, such resource consuming multi-start may not be possible. Hence, we discuss greedy and multi-start separately.
\subsection{Experimental results}

\paragraph{Results of TSP and CVRP.} As shown in \cref{table:tsp_cvrp_100}, \ourmethod{} outperforms the NCO baselines in both the greedy rollout and multi-start settings with the fastest inference speed. Remarkably, \ourmethod{} achieves a $0.95\%$ gap in TSP using the greedy rollout. In the TSP greedy setting, it solves TSP 10,000 instances in a few seconds.

\begin{table}[t]
\centering
\fontsize{8.5}{8.5}\selectfont
\caption{Performance evaluation results for TSP and CVRP. Bold represents the best performances in each task. `-' indicates that the solver does not support the problem. `s' indicates multi-start sampling,  `bs' indicates the beam search. `$\times 5$ for the MDAM indicates the 5 decoder ensemble.}
\begin{tabular}{lccccccc}
\specialrule{1pt}{0pt}{4pt}
\multicolumn{2}{c}{\begin{tabular}{c}\multirow{2}{*}{Method}\end{tabular}}&\multicolumn{3}{c}{TSP ($N = 100$) }&\multicolumn{3}{c}{CVRP ($N = 100$) }\\\cmidrule[0.5pt](lr{0.2em}){3-5} \cmidrule[0.5pt](lr{0.2em}){6-8} 
\multicolumn{2}{c}{}&\multicolumn{1}{c}{Cost $\downarrow$}&\multicolumn{1}{c}{Gap}&\multicolumn{1}{c}{Time}&\multicolumn{1}{c}{Cost $\downarrow$}&\multicolumn{1}{c}{ Gap}&\multicolumn{1}{c}{Time}\\
\specialrule{1.0pt}{2pt}{4pt}

\multicolumn{8}{l}{\textit{Handcrafted Heuristic-based Classical Methods}}\\
\cmidrule[0.5pt](lr{0.1em}){1-8}
Concorde&{Heuristic \cite{concorde}}&{7.76}&0.00$\%$&{3m}&\multicolumn{3}{c}{--}\\
LKH3&{Heuristic \cite{lkh2017} }   &{7.76}&0.00$\%$&{21m} &{15.65}&0.00$\%$&{13h}\\
\cmidrule[0.5pt](lr{0.1em}){1-8}

\multicolumn{8}{l}{\textit{RL-based Deep Constructive Heuristic methods with greedy rollout}}\\
\cmidrule[0.5pt](lr{0.1em}){1-8}

PointerNet $\{greedy.\}$ &{NIPS'15 \cite{pointer,bello2017neural}} & 8.60 & 6.90 $\%$ & {--} & \multicolumn{3}{c}{--}\\
S2V-DQN $\{greedy.\}$ &{NIPS'17 \cite{Khalil}} & 8.31 & 7.03 $\%$ & {--} & \multicolumn{3}{c}{--}\\
RL $\{greedy.\}$ & {NeurIPS'18 \cite{Nazari}} & \multicolumn{3}{c}{--} & 17.23 & 10.12$\%$ & {--}\\
AM $\{greedy.\}$&{ICLR'19 \cite{kool2018attention}}   &{8.12}&{4.53$\%$}&{2s}&16.80&7.34$\%$&{3s}\\

MDAM \{$greedy.\times$ 5\}& {AAAI'21 \cite{ma2021learning}}  &{7.93}&{2.19$$\%$$}&{36s}&16.40& 4.86$$\%$$&{45s}\\

POMO $\{greedy.\}$&{NeurIPS'20 \cite{kwon2020pomo}} &{7.85}&{1.04$\%$}&{2s}&16.26&3.93$\%$&{3s}\\

\textbf{Sym-NCO}  $\{greedy.\}$&{\textit{This work}}  &{\textbf{7.84}}&{\textbf{0.94$\%$}}&{2s}&\textbf{16.10}& \textbf{2.88$\%$}&{3s}\\

\cmidrule[0.5pt](lr{0.1em}){1-8}

\multicolumn{8}{l}{\textit{RL-based Deep Constructive Heuristic methods with multi-start rollout}}\\

\cmidrule[0.5pt](lr{0.1em}){1-8}

Nazari et al. $\{bs. 10\}$  & {NeurIPS'18 \cite{Nazari}} & \multicolumn{3}{c}{--} & 16.96 & 8.39$$\%$$ & {--}\\
AM $\{s. 1280\}$ &{ICLR'19 \cite{kool2018attention}}   &{7.94}&{2.26$$\%$$}&{41m}&16.23&3.72$$\%$$ &{54m}\\
POMO \{s. 100\} &{NeurIPS'20 \cite{kwon2020pomo}} &{7.80}&{0.44$$\%$$}&{13s}&15.90&1.67$$\%$$ &{16s}\\
MDAM \{bs. 30 $\times$ 5\}&{AAAI'21 \cite{ma2021learning}}  &{7.80}&{0.48$$\%$$}&{20m}&16.03& 2.49$$\%$$ &{1h}
\\

\textbf{Sym-NCO} $\{s. 100\}$ &{\textit{This work}} &{\textbf{7.79}}&{\textbf{0.39$$\%$$}}&{13s}&{\textbf{15.87}}& {\textbf{1.46$$\%$$}}&{16s}\\
\specialrule{1pt}{1pt}{4pt}
\end{tabular}
\label{table:tsp_cvrp_100}
\vspace{-2em}
\end{table}

\begin{table}[t]
\centering
\fontsize{8.5}{8.5}\selectfont
\caption{Performance evaluation results for PCTSP and OP. Notations are the same with \cref{table:tsp_cvrp_100}.}
\begin{tabular}{lccccccc}
\specialrule{1pt}{0pt}{4pt}
\multicolumn{2}{c}{\begin{tabular}{c}\multirow{2}{*}{Method}\end{tabular}}&\multicolumn{3}{c}{PCTSP ($N = 100$) }&\multicolumn{3}{c}{OP ($N = 100$) }\\\cmidrule[0.5pt](lr{0.2em}){3-5} \cmidrule[0.5pt](lr{0.2em}){6-8} 
\multicolumn{2}{c}{}&\multicolumn{1}{c}{Cost $\downarrow$}&\multicolumn{1}{c}{Gap}&\multicolumn{1}{c}{Time}&\multicolumn{1}{c}{Obj $\uparrow$}&\multicolumn{1}{c}{ Gap}&\multicolumn{1}{c}{Time}\\
\specialrule{1.0pt}{2pt}{4pt}

\multicolumn{8}{l}{\textit{Handcrafted Heuristic-based Classical Methods}}\\
\cmidrule[0.5pt](lr{0.1em}){1-8}
ILS C++ & Heuristic \cite{kool2018attention} &{5.98}&0.00$\%$&{12h}&\multicolumn{3}{c}{--} \\
Compass & Heuristic \cite{kobeaga2018efficient} &\multicolumn{3}{c}{--}&33.19&0.00$\%$&15m \\
\cmidrule[0.5pt](lr{0.1em}){1-8}

\multicolumn{8}{l}{\textit{RL-based Deep Constructive Heuristic methods with greedy rollout (zero-shot inference)}}\\

\cmidrule[0.5pt](lr{0.1em}){1-8}

AM \{$greedy$.\}&{ICLR'19 \cite{kool2018attention}}   &{6.25}&{4.46$\%$}&{2s}&31.62&4.75$\%$&{2s}\\

MDAM \{$greedy$.$\times$ 5\}&{AAAI'21 \cite{ma2021learning}} &{6.17}&{3.13$\%$}&{34s}&32.32& 2.61$\%$&{32s}\\

\textbf{Sym-NCO}  \{$greedy$.\}&{\textit{This work}}  &{\textbf{6.05}}&{\textbf{1.23\%}}&{2s}&\textbf{32.51}&\textbf{2.03\%} &2s\\
 
\cmidrule[0.5pt](lr{0.1em}){1-8}

\multicolumn{8}{l}{\textit{RL-based Deep Constructive Heuristic methods with multi-start rollout (Post-processing)}}\\

\cmidrule[0.5pt](lr{0.1em}){1-8}

AM \{s. 1280\}&{ICLR'19 \cite{kool2018attention}}   &{6.08}&{1.67$\%$}&{27m}&32.68&1.55$\%$&{ 25m}\\

MDAM \{bs. 30$\times$ 5\}&{AAAI'21 \cite{ma2021learning}}  &{6.07}&{1.46$\%$}&{16m}&32.91& 0.84$\%$&{14m}\\
\textbf{Sym-NCO}  \{s. 200\}&{\textit{This work}}  &{\textbf{5.98}}&{\textbf{-0.02\%}}&{3m}&{\textbf{33.04}}& {\textbf{0.45\%}}&{3m}\\
\specialrule{1pt}{1pt}{4pt}

\end{tabular}
\label{table:pctsp_op_100}
\vspace{-1em}
\end{table}


\paragraph{Results of PCTSP and OP.} As shown in \cref{table:pctsp_op_100}, \ourmethod{} outperforms the NCO baselines in both the greedy rollout and multi-start settings. In the multi-start setting, \ourmethod{} outperforms the classical PCTSP baseline (i.e., ILS) with the $\frac{43200}{180} \approx 240 \times$ faster speed. 

\begin{wraptable}{r}{0.2\textwidth}
\centering
\begin{tabular}{lc}
\hline
{} & Gap \\
\hline
POMO & 1.87\% \\
\ourmethod{} & \textbf{1.62\%} \\
\hline
\end{tabular}
\caption{Optimality gap on TSPLIB}
\label{table:tsplib_results}
\vspace{-1.6em}
\end{wraptable}

\paragraph{Results of the real-world TSP.} We evaluate POMO and \ourmethod{} on TSPLib \cite{reinelt1991tsplib}. \cref{table:tsplib_results} shows that \ourmethod{} outperforms POMO. Please refer to \cref{append:TSPLIB} for the full benchmark results.

\begin{figure}[t]
     \centering
     \begin{subfigure}[b]{0.31\textwidth}
         \centering
         \includegraphics[width=\textwidth]{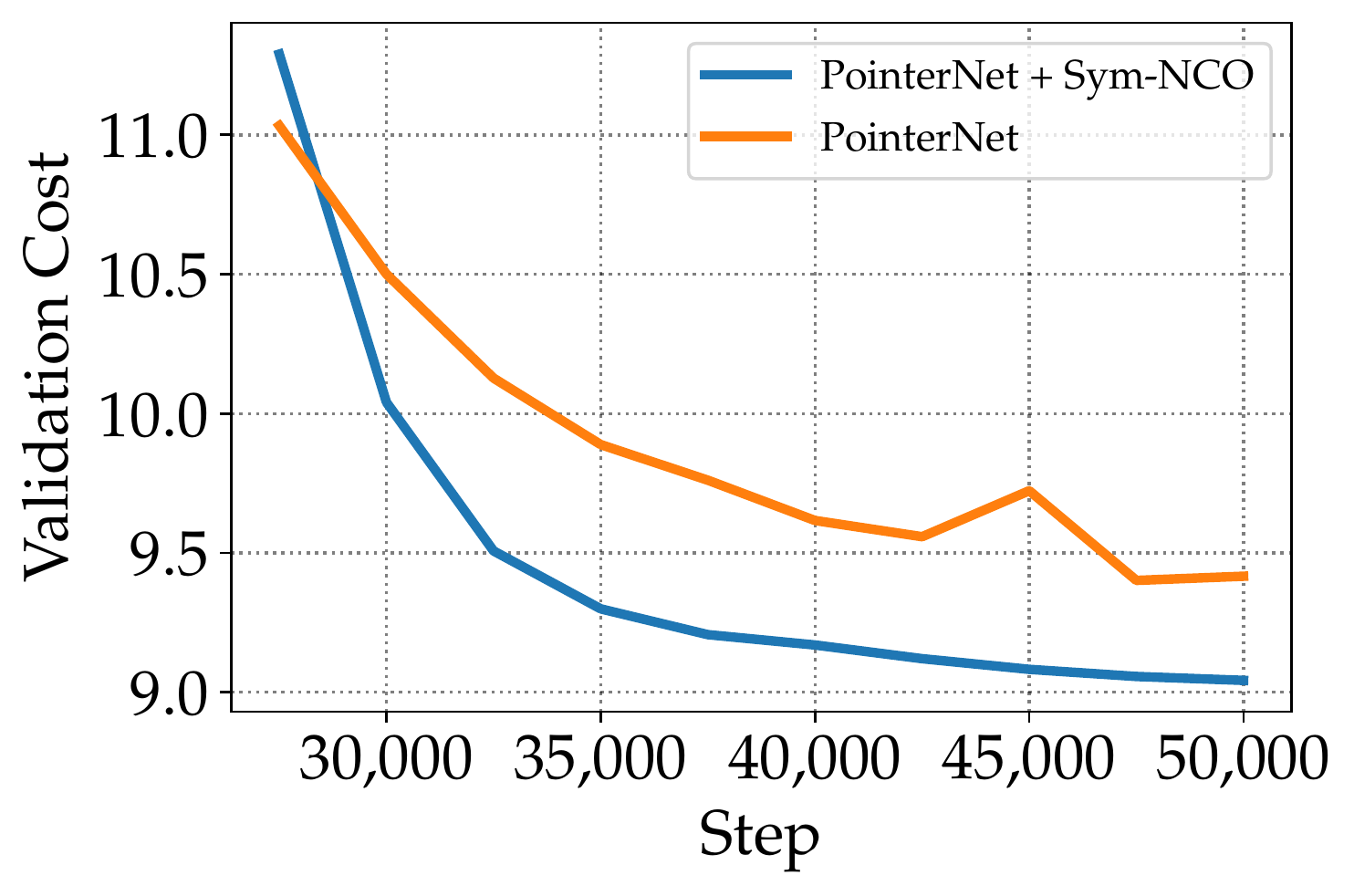}
         \caption{PointerNet}
         \label{fig:pointernet-tsp100}
     \end{subfigure}
     \hfill
     \begin{subfigure}[b]{0.31\textwidth}
         \centering
        \includegraphics[width=\textwidth]{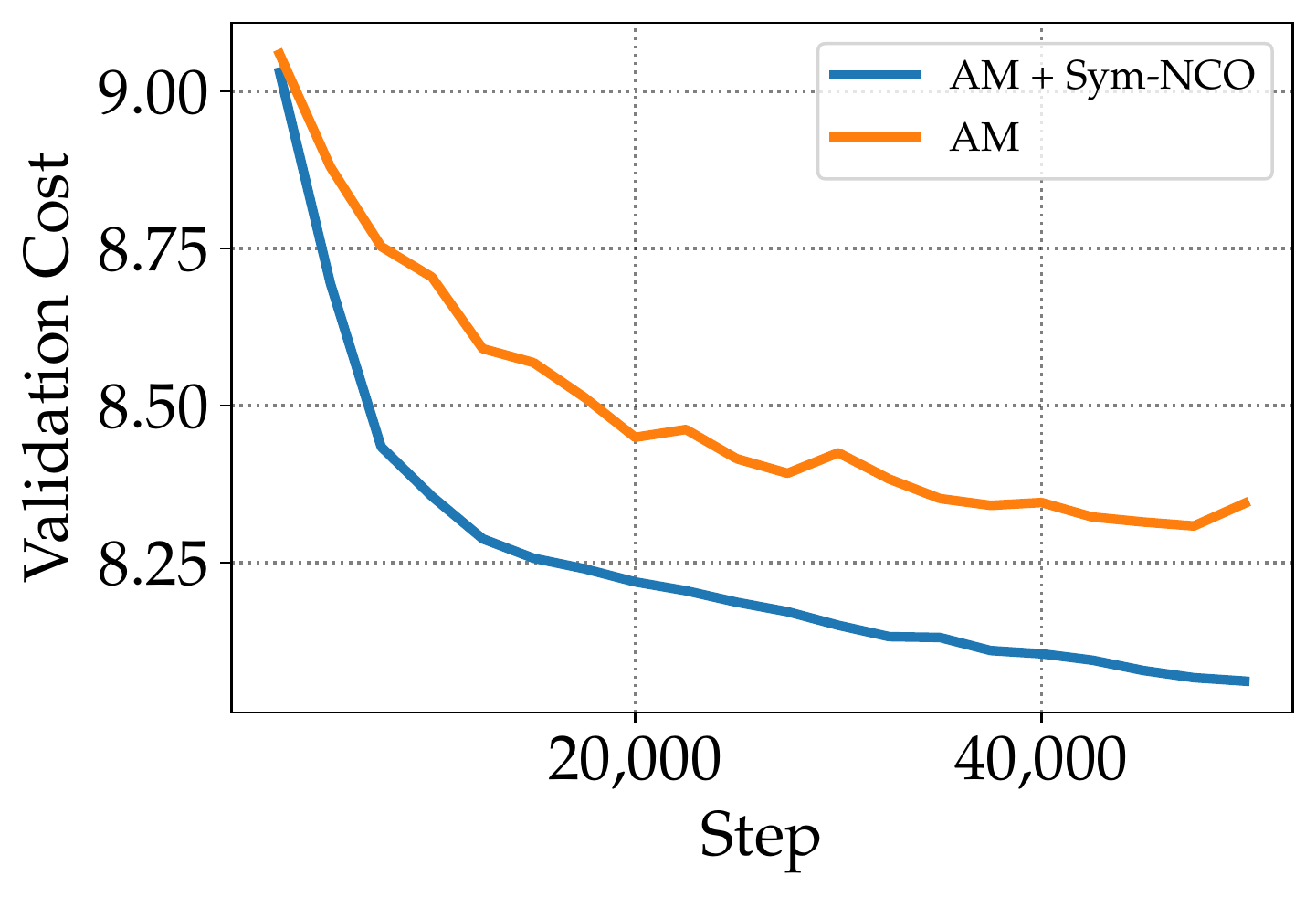}
         \caption{AM}
         \label{fig:am-tsp100}
     \end{subfigure}
     \hfill
          \begin{subfigure}[b]{0.31\textwidth}
         \centering
        \includegraphics[width=\textwidth]{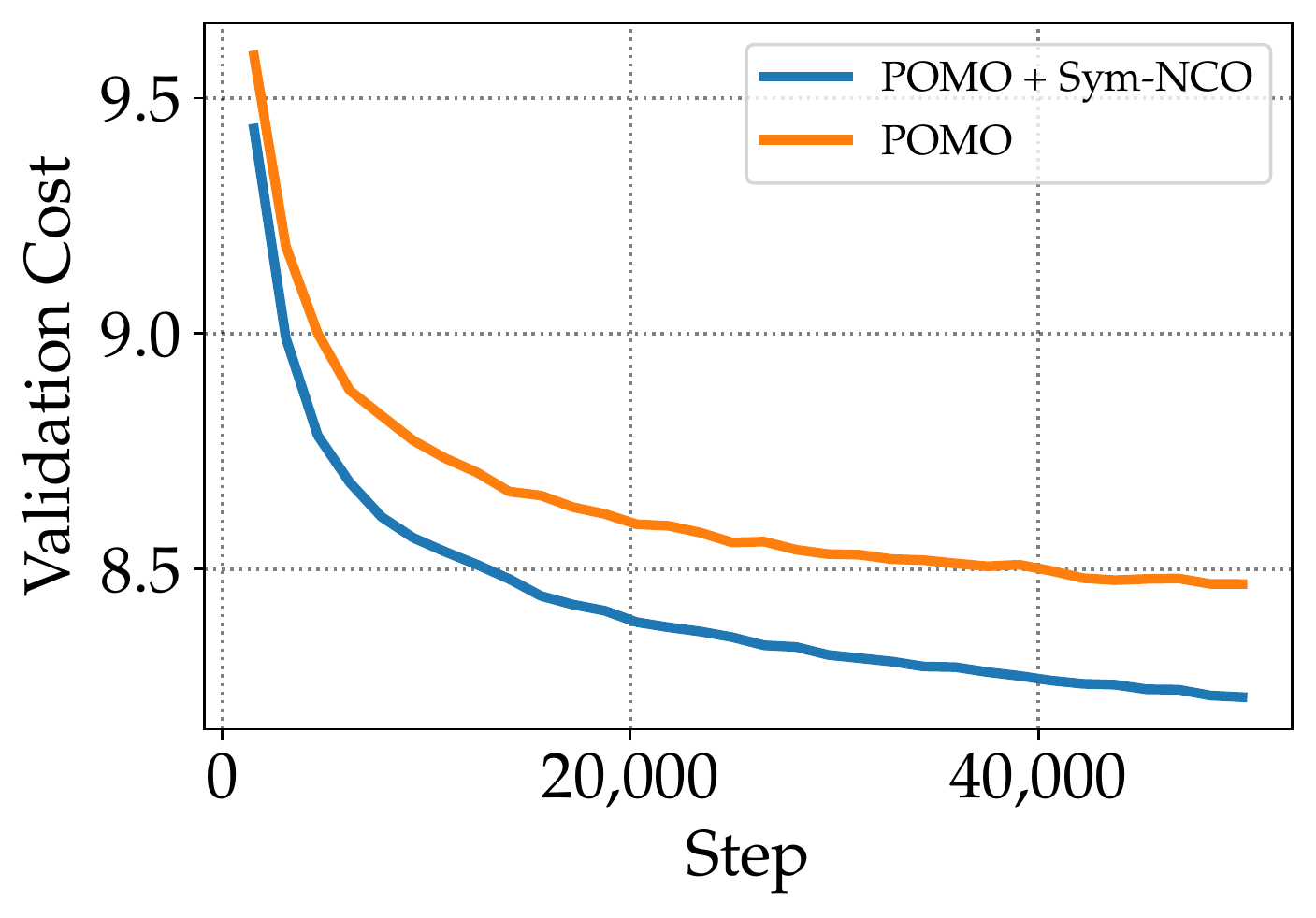}
         \caption{POMO}
     \end{subfigure}
    \caption{The applications of \ourmethod{} to DRL-NCO methods in TSP ($N=100)$}
    \label{figure:drl-nco-with-sym-nco}
    \vspace{-1.6em}
\end{figure}

\paragraph{Application to various DRL-NCO methods.} As discussed in \cref{sec:sym-nco}, \ourmethod{} can be applied to various various DRL-NCO methods. We validate that \ourmethod{} significantly improves the existing DRL-NCO methods as shown in \cref{figure:drl-nco-with-sym-nco}.


\begin{figure}[t]
     \centering
     \begin{subfigure}[b]{0.245\textwidth}
         \centering
         \includegraphics[width=\textwidth]{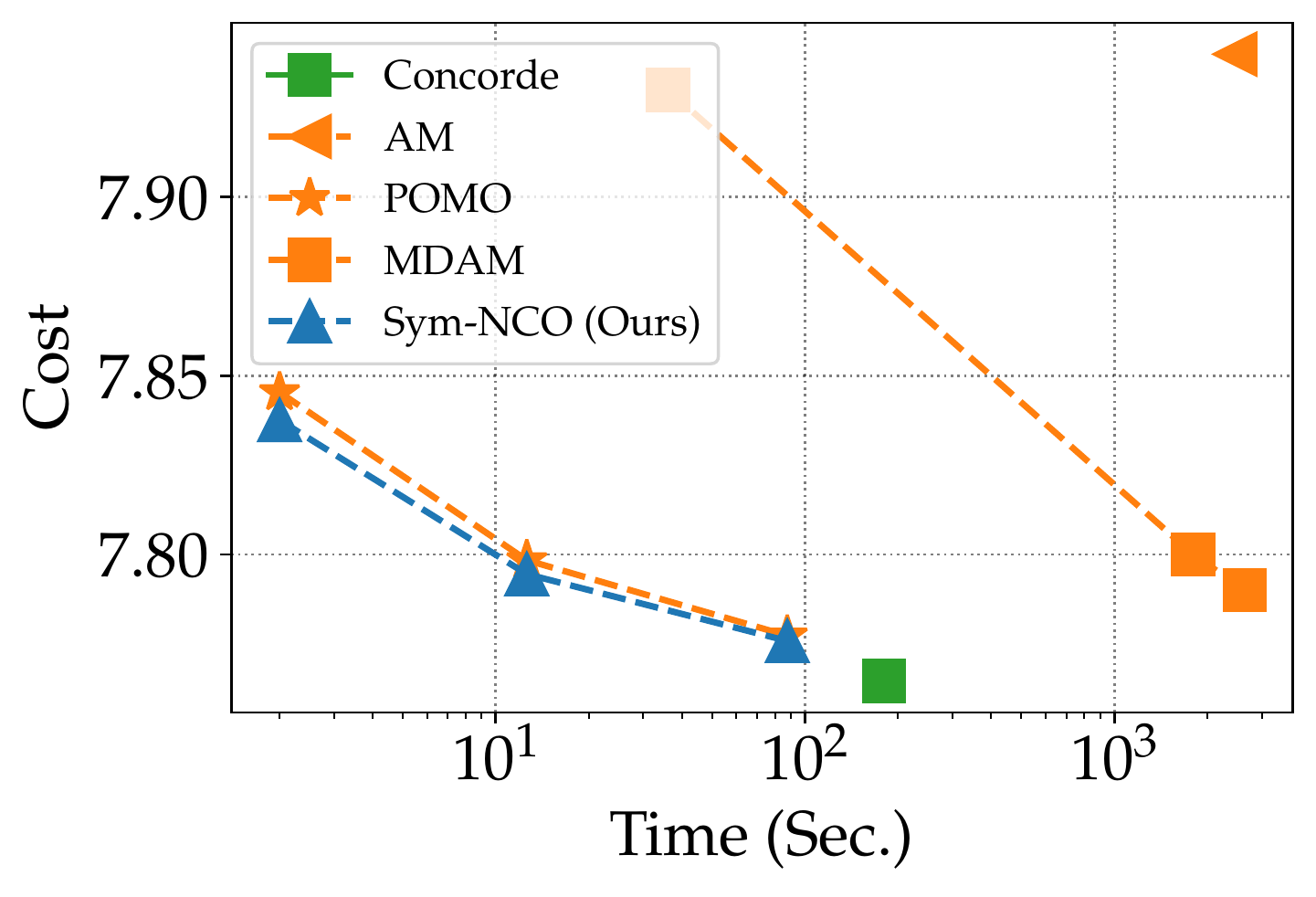}
         \caption{TSP ($N=100$)}
         \label{figure:time_perf_TSP100}
     \end{subfigure}
     \hfill
     \begin{subfigure}[b]{0.245\textwidth}
         \centering
         \includegraphics[width=\textwidth]{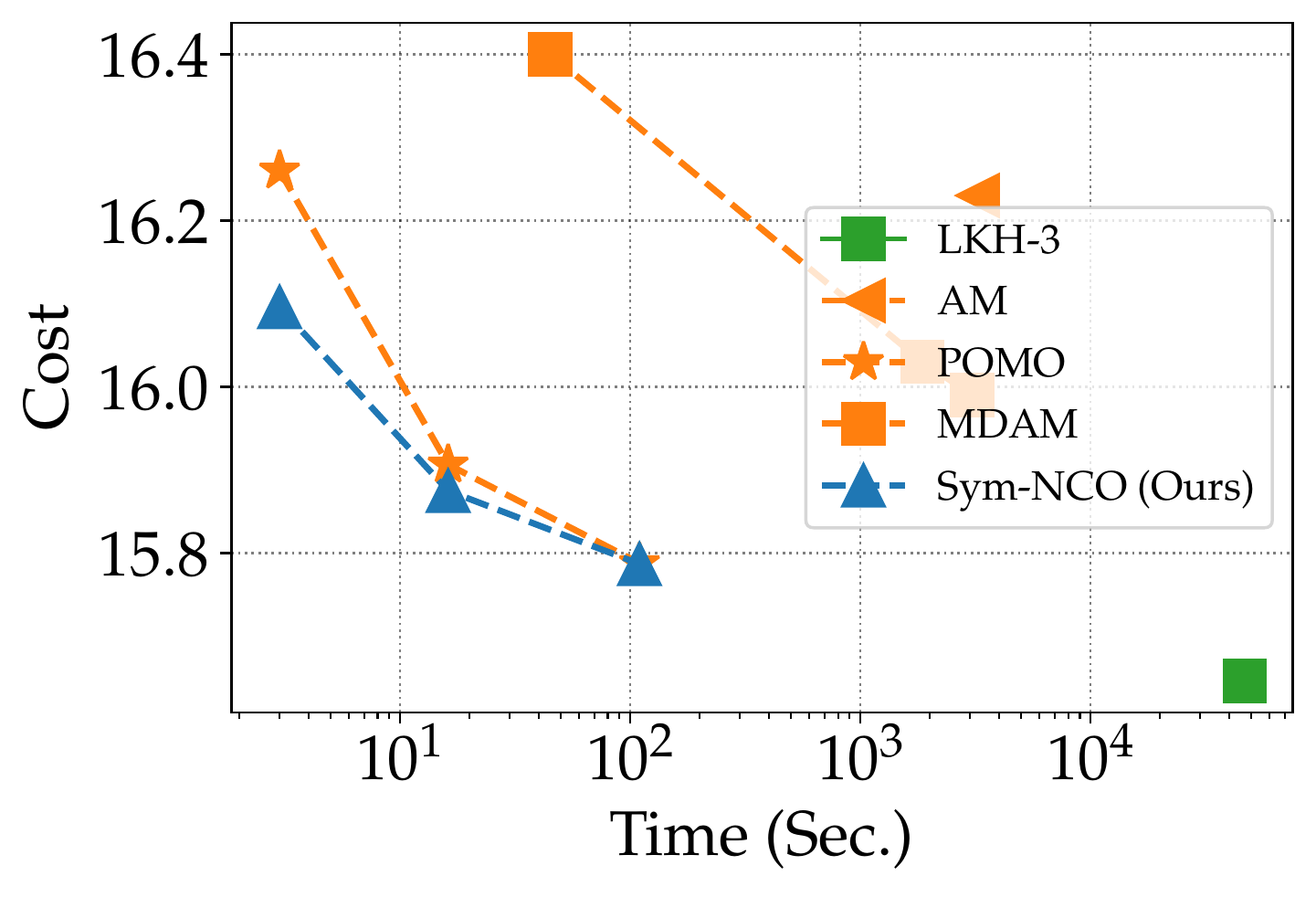}
         \caption{CVRP ($N=100$)}
         \label{figure:time_perf_CVRP100}
     \end{subfigure}
     \hfill
          \begin{subfigure}[b]{0.245\textwidth}
         \centering
         \includegraphics[width=\textwidth]{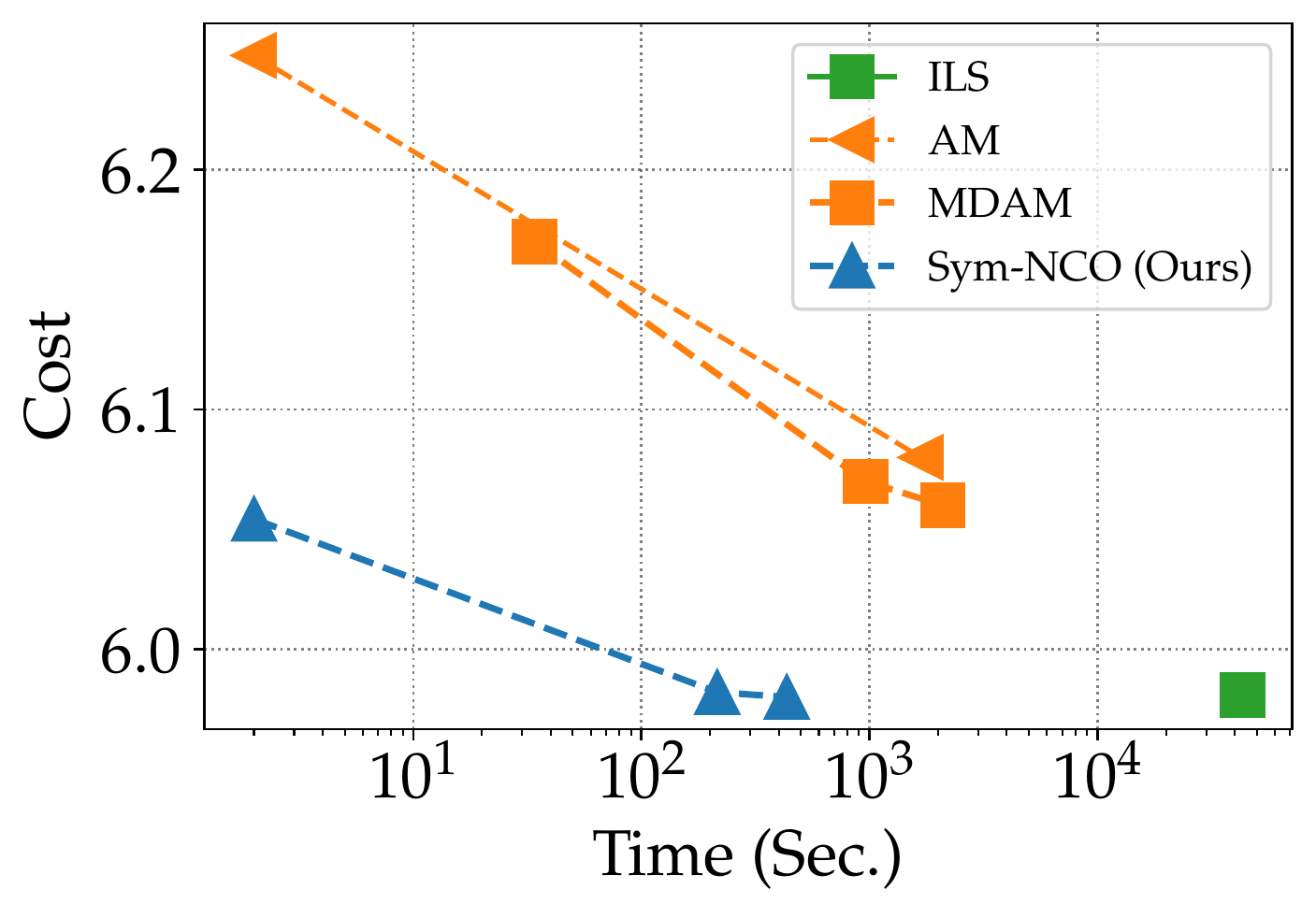}
         \caption{PCTSP ($N=100$)}
         \label{figure:time_perf_PCTSP100}
     \end{subfigure}
     \hfill
          \begin{subfigure}[b]{0.245\textwidth}
         \centering
         \includegraphics[width=\textwidth]{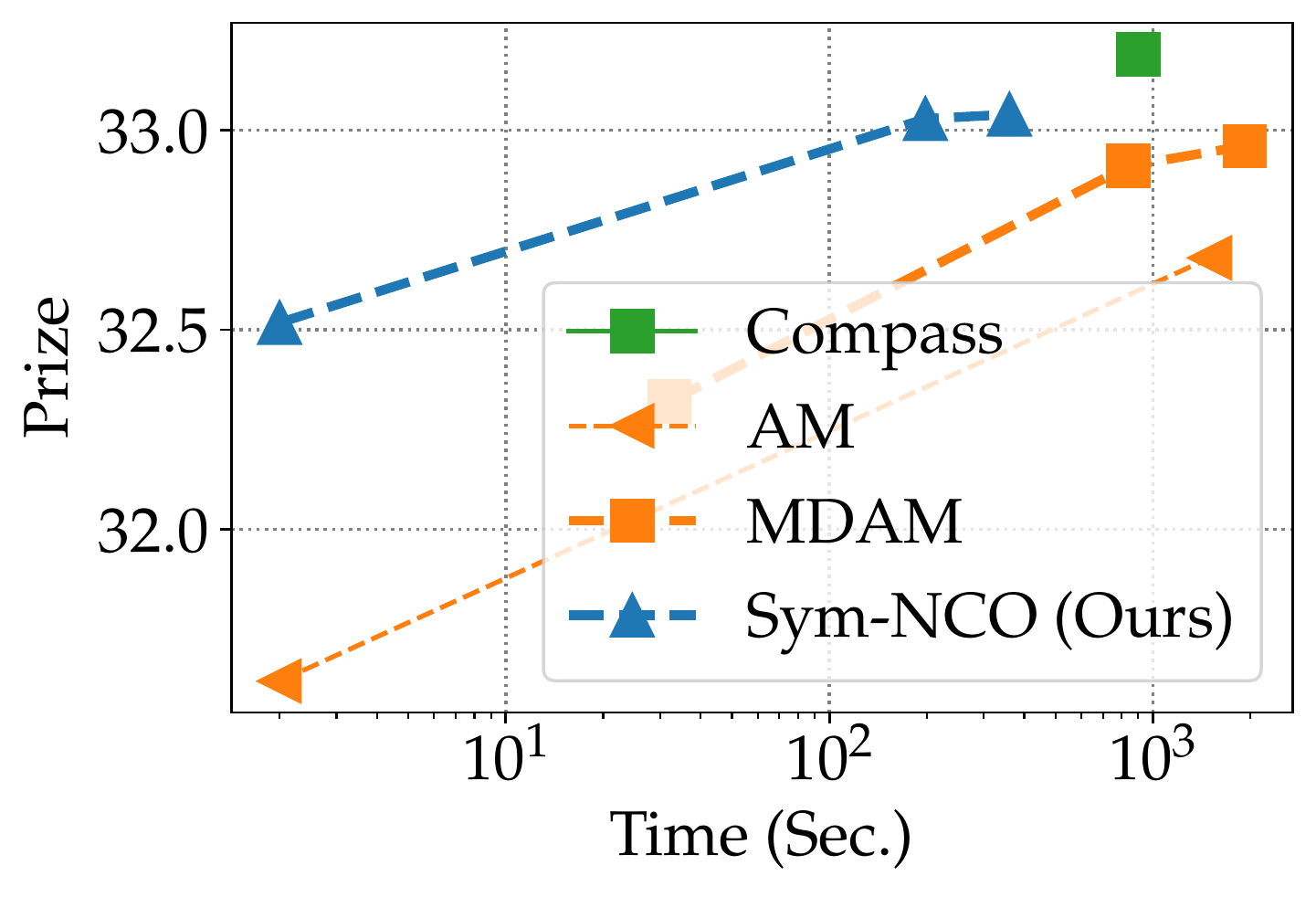}
         \caption{OP ($N=100$)}
         \label{figure:time_perf_OP100}
     \end{subfigure}
    \caption{Time vs. cost plots. \textcolor{ForestGreen}{Green}, \textcolor{Orange}{orange}, and \textcolor{NavyBlue}{blue} colored lines visualize the results of \textcolor{ForestGreen}{hand-craft heuristics}, \textcolor{Orange}{neural baselines}, and \textcolor{Blue}{\ourmethod{}}, respectively.  For OP (d), higher y-axis values are better.}
    \label{figure:time_perf_analysis}
    \vspace{-8pt}
\end{figure}

\paragraph{Time-performance analysis for multi-starts.} Multi-starts is a common method that improves the solution qualities while requiring a longer time budget. We use the rotation augments \cite{kwon2020pomo} to produce multiple inputs (i.e., starts). As shown in \cref{figure:time_perf_analysis}, \ourmethod{} achieves the Pareto frontier for all benchmark datasets. In other words, \ourmethod{} exhibits the best solution quality among the baselines within the given time consumption.




\section{Discussion}
\begin{figure}[t]
     \centering
     \begin{subfigure}[t]{0.32\textwidth}
         \centering
         \includegraphics[width=\textwidth]{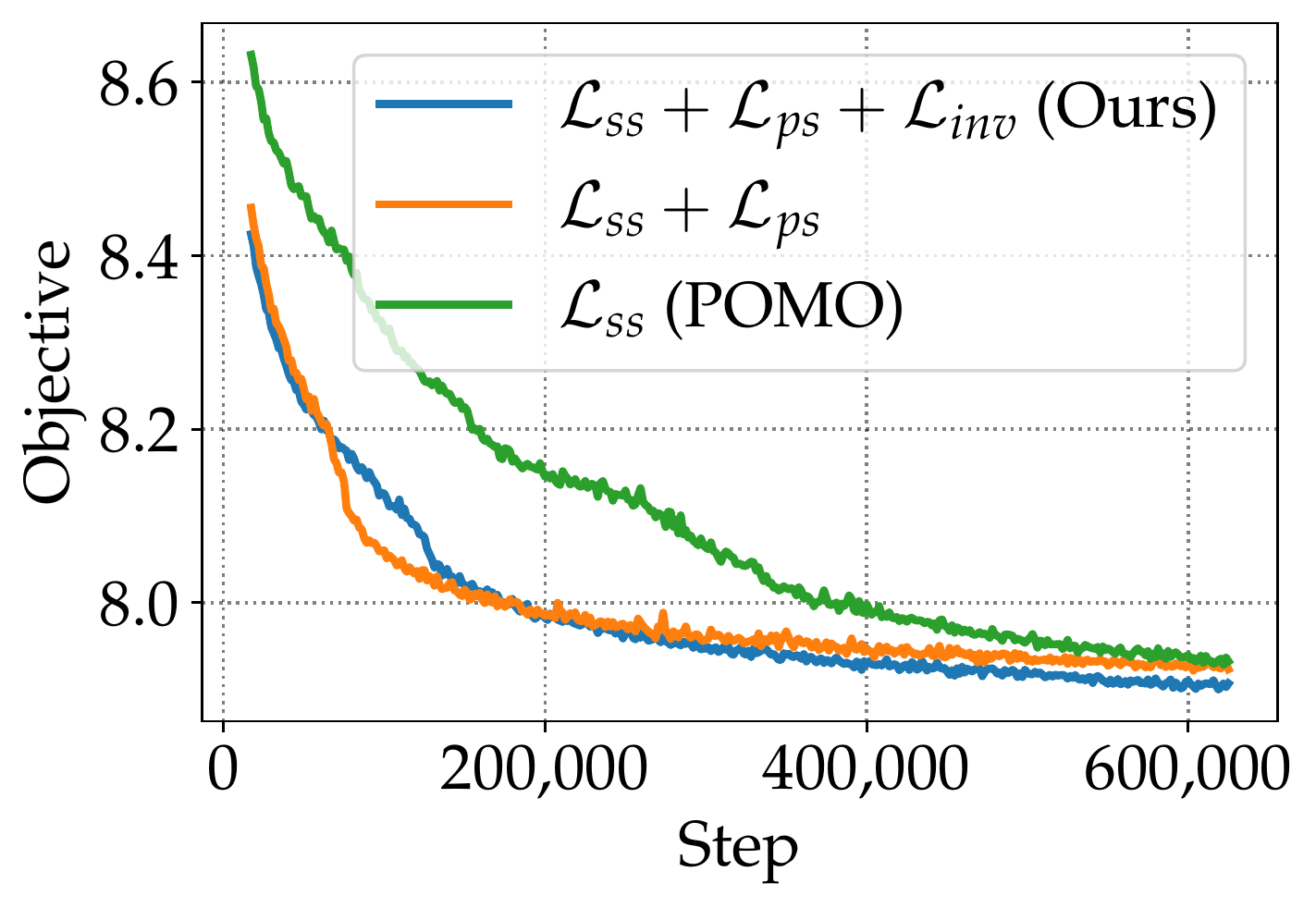}
         \caption{Cost curves of the loss designs}
         \label{fig:loss_design_cost_plot}
     \end{subfigure}
     \hfill
     \begin{subfigure}[t]{0.32\textwidth}
        \centering
        \includegraphics[width=\textwidth]{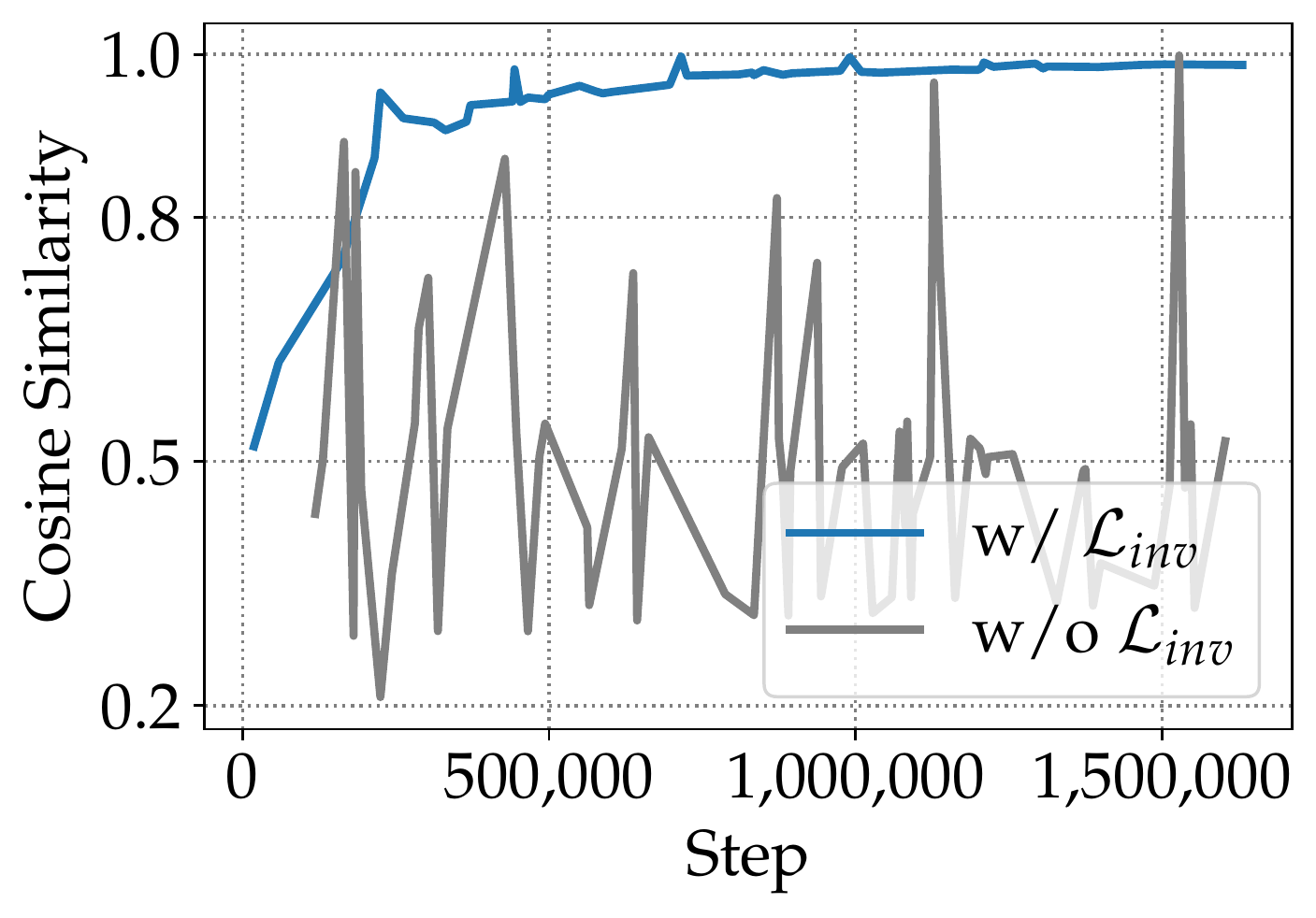}
        \caption{Cosine similarity curves}
        \label{fig:cos_sim_plot}
     \end{subfigure}
        \begin{subfigure}[t]{0.32\textwidth}
        \centering
        \includegraphics[width=\textwidth]{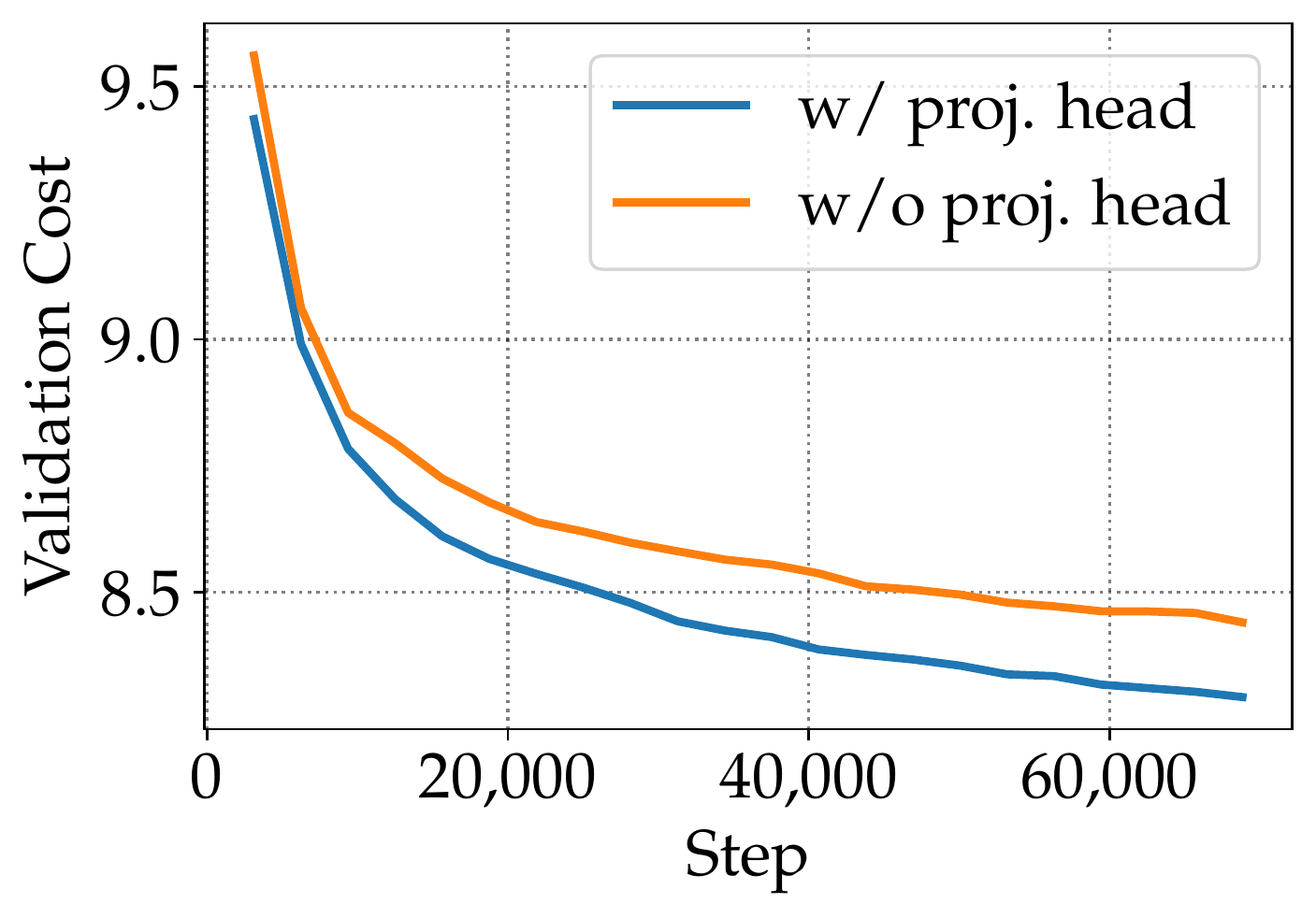}
        \caption{Projection head ablation results}
        \label{fig:proj_head_plot}
     \end{subfigure}
    \caption{Loss design ablation results (a) Effect of loss components to the costs, (b) Cosine similarity curves of the models with and with $\mathcal{L}_{\text{inv}}$, (c) Costs of the models with and without $g(\cdot)$.}
    \vspace{-1em}
\end{figure}

\subsection{Discussion of Regularization based Symmetricity Learning}
\label{subsec:hard_const}

\begin{figure}[t]
    \centering
    \begin{subfigure}[b]{0.45\textwidth}
        \centering
        \includegraphics[width=\textwidth]{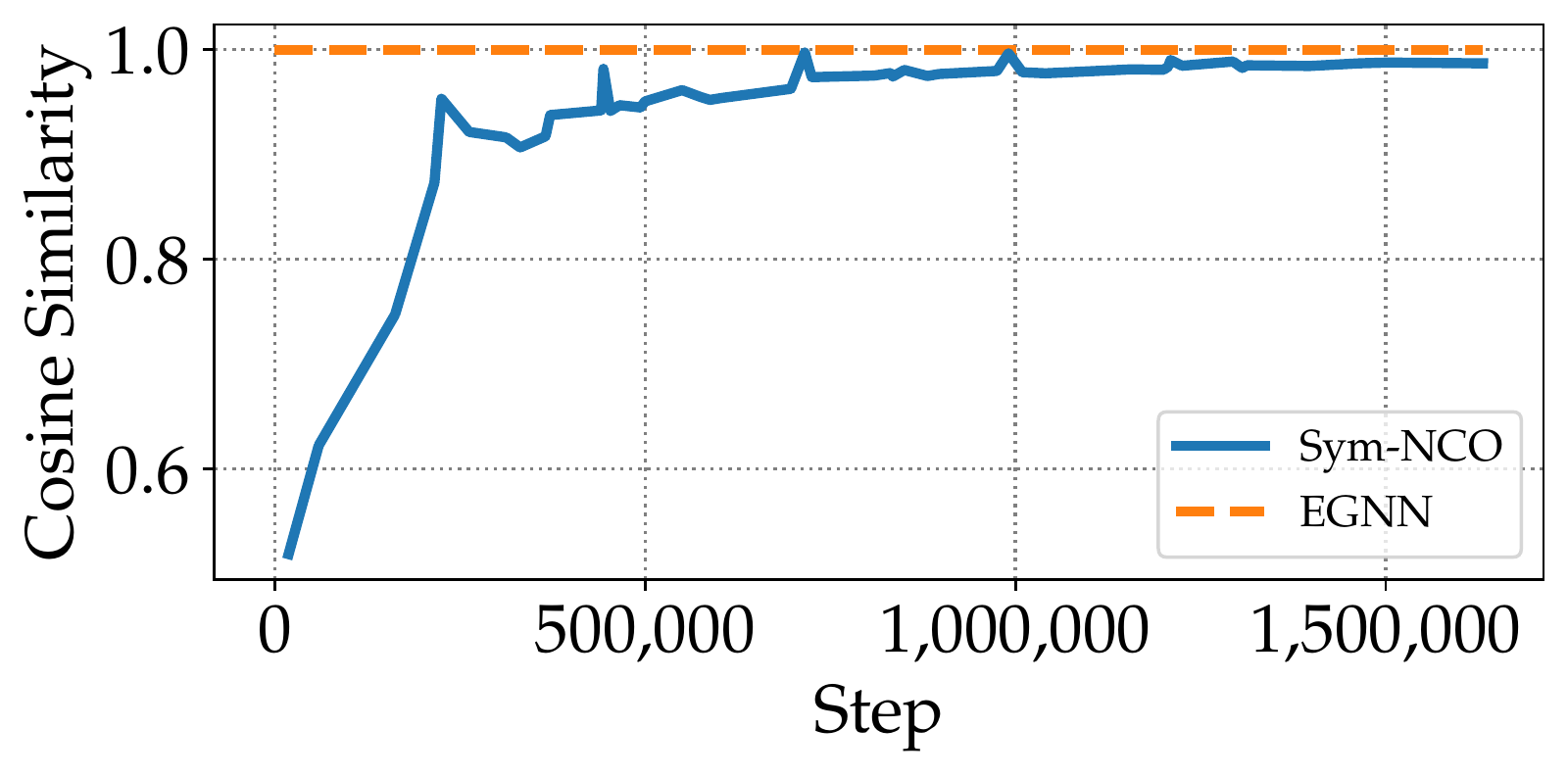}
        \caption{Cosine similarity curves}
        \label{figure:cos_sim_egnn_sym}
    \end{subfigure}
    \hfill
    \begin{subfigure}[b]{0.45\textwidth}
        \centering
        \includegraphics[width=\textwidth]{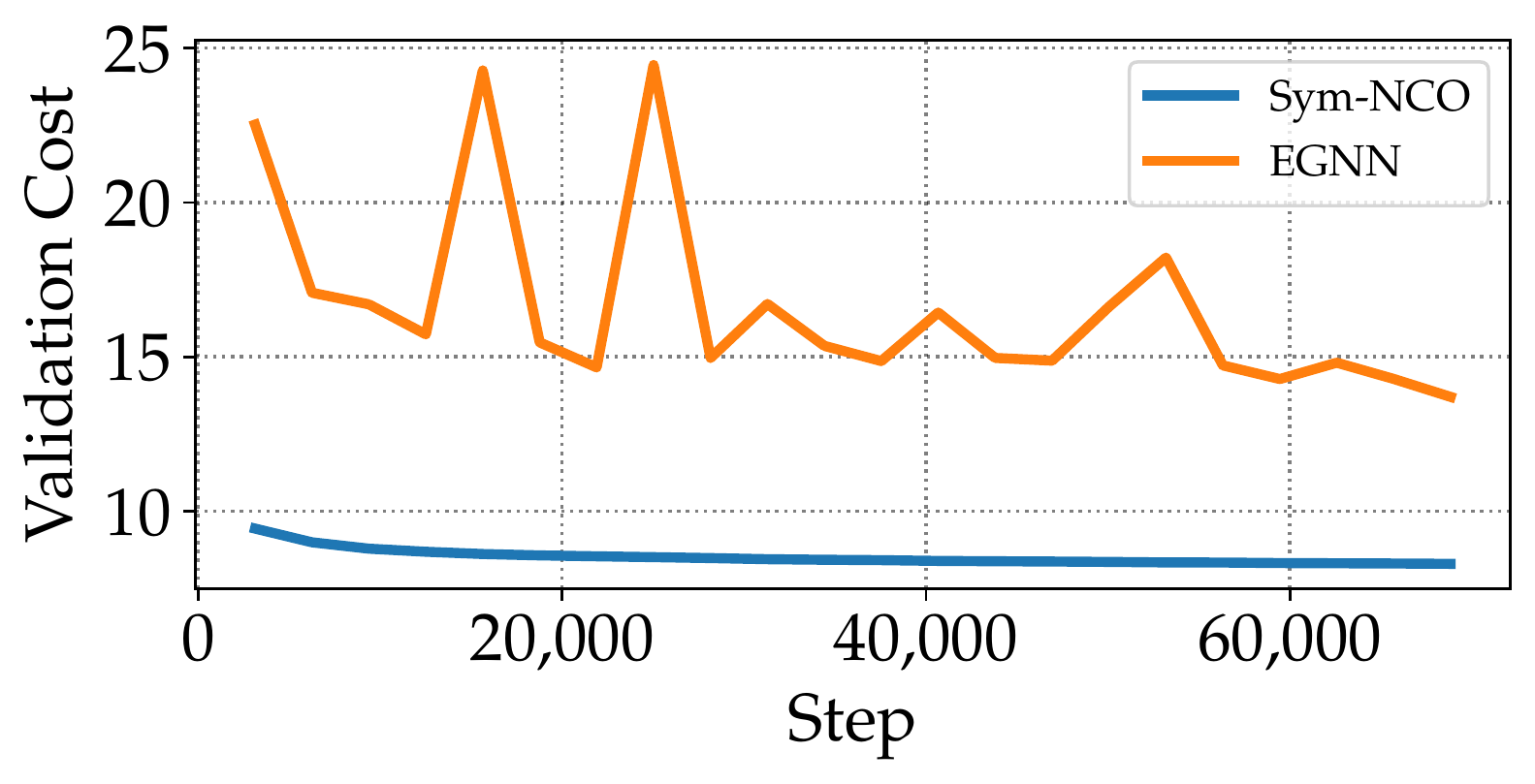}
        \caption{Cost curves}
        \label{figure:perf_egnn_sym}
    \end{subfigure}
    \caption{Comparisons of \ourmethod{} and EGNN}
\end{figure}



\paragraph{Ablation Study of $\mathcal{L}_{\text{inv}}$.} As shown in \cref{fig:cos_sim_plot}, $\mathcal{L}_{\text{inv}}$ increases the cosine similarity of the projected representation (i.e., $g(h)$). We can conclude that $\mathcal{L}_{inv}$ contributes to the performance improvements (see \cref{fig:loss_design_cost_plot}). We further verify that imposing similarity on $h$ degrades the performance as demonstrated in \cref{fig:proj_head_plot}. This proves the importance of maintaining the expression power of the encoder as we mentioned in \cref{subsec:invariant}. 

\paragraph{Comparison with EGNN.} 

EGNN \cite{satorras2021n} provably guarantees to process coordinate graph with rotational equivalency. Also, EGNN empirically verified its' high performance in point cloud tasks. Therefore, we implemented a simple EGNN-based CO policy (simply termed EGNN in this paper), to check the feasibility of CO. We leverage six EGNN layers with 128 hidden dimensions, to replace the POMO encoder where the POMO decoder is unchanged. 

In the experimental results, we observed that EGNN significantly underperforms \ourmethod{} and fails to converge as shown \cref{figure:perf_egnn_sym}. This is because the euclidian CO has a fully connected input graph containing informative coordinates, and we believe the equivariant neural network should be carefully crafted to consider such unique input graph structures of CO tasks. On the other hand, \ourmethod{} could leverage the existing powerful NCO model without fine modification of the neural network. These numerical results conform well with our hypothesis that equivariance is necessary but not sufficient for obtaining the optimal parameters for the solver. Therefore, we expect our approach can be simply extended to other domains requiring symmetricity and geometricity as positioned to leverage the existing legacy of powerful non-equivariant neural networks.  



\subsection{Limitations \& future directions}
\label{sec:limit}
\paragraph{Extended problem symmetricities.} In this work, we employ the rotational symmetricity (\cref{thm:rot_sym}) as the problem symmetricity. However, for some COPs, different problem symmetricities, such as scaling and translating $P$, can also be considered. Employing these additional symmetricities may further enhance the performance of \ourmethod{}. We leave this for future research.\vspace{-2mm}
\paragraph{Large scale adaptation.} Large scale applicability is essential to NCO.  In this work we simply present scale adaptation capability of \ourmethod{} using the effective active search (EAS) \cite{hottung2021efficient}; see \cref{large_scale}.
We expect curriculum- \cite{lisicki2020evaluating}, and meta-learning approaches may improve the generalizability of NCO to larger-sized problems. 

\paragraph{Extension to the graph COP.} This work finds the problem symmetricity that is universally applicable for \textit{Euclidean} COPs. However, some COPs are defined in non-Euclidean spaces such as asymmetric TSP. Furthermore, there are also a bunch of existing neural combinatorial optimization models that can solve graph COP \cite{barrett2020exploratory,drori2020learning,barrett2022learning,math8020298,abe2019solving,li2018combinatorial}, where we can improve with a symmetricity regularization scheme. We also leave finding the universal symmetricities of non-Euclidean COPs and applying them to existing graph COP models for future research.

\subsection{Social Impacts}
Design automation through NCO research affects various industries including logistics and transportation industries. From a negative perspective, this automation process may have some concerns to lead to unemployment in certain jobs. However, automation of logistics, transportation, and design automation can increase the efficiency of industries, reducing CO2 emissions (by reducing total tour length) and creating new industries and jobs.



\section*{Acknowledgments and Disclosure of Funding}

We thank Jiwoo Son, Hyeonah Kim, Haeyeon Rachel Kim, Fangying Chen, and anonymous reviews for proving helpful feedback for preparing our manuscripts. This work was supported by a grant of the KAIST-KT joint research project through AI2XL Laboratory, Institute of convergence Technology, funded by KT [Project No. G01210696, Development of Multi-Agent Reinforcement Learning Algorithm for Efficient Operation of Complex Distributed Systems].


\bibliography{main}

\newpage

\begin{enumerate}

\item For all authors...
\begin{enumerate}
  \item Do the main claims made in the abstract and introduction accurately reflect the paper's contributions and scope?
    \answerYes{} 
    
  \item Have you read the ethics review guidelines and ensured that your paper conforms to them?
    \answerYes{}
    
  \item Did you discuss any potential negative societal impacts of your work?
    \answerNA{}
    
  \item Did you describe the limitations of your work?
    \answerYes{} See \cref{sec:limit}
    
\end{enumerate}

\item If you are including theoretical results...
\begin{enumerate}
  \item Did you state the full set of assumptions of all theoretical results?
    \answerNA{}
    
	\item Did you include complete proofs of all theoretical results?
    \answerNA{}
    
\end{enumerate}

\item If you ran experiments...
\begin{enumerate}
  \item Did you include the code, data, and instructions needed to reproduce the main experimental results (either in the supplemental material or as a URL)?
    \answerYes{} Source code will be available after decision is made.
    
  \item Did you specify all the training details (e.g., data splits, hyperparameters, how they were chosen)?
    \answerYes{} See \cref{sec:exp_setting}.
    
	\item Did you report error bars (e.g., with respect to the random seed after running experiments multiple times)?
    \answerNA{}
    
	\item Did you include the amount of compute and the type of resources used (e.g., type of GPUs, internal cluster, or cloud provider)?
    \answerYes{} See \cref{sec:implement}.
    
\end{enumerate}

\item If you are using existing assets (e.g., code, data, models) or curating/releasing new assets...
\begin{enumerate}
  \item If your work uses existing assets, did you cite the creators?
    \answerYes{}
    
  \item Did you mention the license of the assets?
    \answerNA{}
    
  \item Did you include any new assets either in the supplemental material or as a URL?
    \answerNA{}
    
  \item Did you discuss whether and how consent was obtained from people whose data you're using/curating?
    \answerNA{}
    
  \item Did you discuss whether the data you are using/curating contains personally identifiable information or offensive content?
    \answerNA{}
    
\end{enumerate}

\item If you used crowdsourcing or conducted research with human subjects...
\begin{enumerate}
  \item Did you include the full text of instructions given to participants and screenshots, if applicable?
    \answerNA{}
    
  \item Did you describe any potential participant risks, with links to Institutional Review Board (IRB) approvals, if applicable?
    \answerNA{}
    
  \item Did you include the estimated hourly wage paid to participants and the total amount spent on participant compensation?
    \answerNA{}
    
\end{enumerate}

\end{enumerate}

\appendix
\newpage

\section{Proof of \cref{thm:rot_sym}}
\label{append: proof}
In this section, we prove the \cref{thm:rot_sym}, which states a problem $\boldsymbol{P}$ and its' orthogonal transformed problem $Q(\boldsymbol{P})=\{\{Qx_i\}_{i=1}^{N},\boldsymbol{f}\}$ have identical optimal solutions if $Q$ is orthogonal matrix: $QQ^{T} = Q^{T}Q = I$.

As we mentioned in \cref{subsec:sym-co-mdp}, reward $R$ is a function of $ \boldsymbol{a}_{1:T}$ (solution sequences), \\$||x_i - x_j||_{i,j \in \{1,...N\}}$ (relative distances) and $\boldsymbol{f}$ (nodes features).  

For simple notation, let denote $||x_i - x_j||_{i,j \in \{1,...N\}}$ as $||x_i - x_j||$. And Let $R^{*}(\boldsymbol{P})$ is optimal value of problem $\boldsymbol{P}$: i.e. 

\begin{equation*}
R^{*}(\boldsymbol{P}) = R(\sol^{*};\boldsymbol{P}) = R\left(\sol^{*};\{||x_i - x_j||, \boldsymbol{f}\}\right)    
\end{equation*}

Where $\pi^*$ is an optimal solution of problem $\boldsymbol{P}$. Then the optimal value of transformed problem $Q(\boldsymbol{P})$,  $R^{*}(Q(\boldsymbol{P}))$ is invariant:

\begin{align*}
    R^{*}(Q(\boldsymbol{P})) &= R(\sol^{*};Q(\boldsymbol{P}))\\ &= R\left(\sol^{*};\{||Qx_i - Qx_j||, \boldsymbol{f}\}\right)\\ &= R\left(\sol^{*};\{\sqrt{(Qx_i - Qx_j)^{T}(Qx_i - Qx_j)}, \boldsymbol{f}\}\right)\\ &= R\left(\sol^{*};\{\sqrt{(x_i - x_j)^{T}Q^{T}Q(x_i - x_j)}, \boldsymbol{f}\}\right)\\ &= R\left(\sol^{*};\{\sqrt{(x_i - x_j)^{T}I(x_i - x_j)}, \boldsymbol{f}\}\right)\\ &= R\left(\sol^{*};\{||x_i - x_j||, \boldsymbol{f}\}\right) = R(\sol^{*};\boldsymbol{P}) = R^{*}(\boldsymbol{P})
\end{align*}

Therefore, problem transformation of orthogonal matrix $Q$ does not change the optimal value.

Then, the remaining proof is to show $Q(\boldsymbol{P})$ has an identical solution set with $\boldsymbol{P}$. 

Let optimal solution set $\Pi^{*}(P) = \{\sol^{i}(\boldsymbol{P})\}_{i=1}^{M}$, where $\sol^{i}(\boldsymbol{P})$ indicates optimal solution of $\boldsymbol{P}$ and $M$ is the number of heterogeneous optimal solution. 

For any $\sol^{i}(Q(\boldsymbol{P})) \in \Pi^{*}(Q(\boldsymbol{P}))$, they have same optimal value with $\boldsymbol{P}$:
\begin{align*}
    R(\sol^{i}(Q(\boldsymbol{P}));Q(\boldsymbol{P})) = R^{*}(Q(\boldsymbol{P})) = R^{*}(\boldsymbol{P}) 
\end{align*}

Thus, $\sol^{i}(Q(\boldsymbol{P})) \in \Pi^{*}(P)$. 

Conversely, For any $\sol^{i}(\boldsymbol{P}) \in \Pi^{*}(\boldsymbol{P})$, they have sample optimal value with $Q(\boldsymbol{P})$: 

\begin{align*}
    R(\sol^{i}(\boldsymbol{P});\boldsymbol{P}) = R^{*}(\boldsymbol{P}) = R^{*}(Q(\boldsymbol{P}))
\end{align*}

Thus, $\sol^{i}(\boldsymbol{P}) \in \Pi^{*}(Q(\boldsymbol{P}))$. 

\begin{center}
    Therefore, $\Pi^{*}(\boldsymbol{P}) = \Pi^{*}(Q(\boldsymbol{P}))$, i.e., $\boldsymbol{P} \sym Q(\boldsymbol{P})$.
\end{center}

\clearpage

\section{Implementation of Baselines}
\label{append: baseline}
We directly reproduce competitive DRL-NCO methods: POMO \cite{kwon2020pomo} and AM \cite{kool2018attention} and PointerNet \cite{pointer,bello2017neural}. 

\textbf{PointerNet.} The PointerNet is early work of DRL-NCO using LSTM-based encoder-decoder architecture trained with actor-critic manner. We follow the instruction of open source code \footnote{https://github.com/wouterkool/attention-learn-to-route} by \cite{kool2018attention} following hyperparmeters. 

\begin{table}[h]
    \centering
    \begin{tabular}{lc}
        \hline
        REINFORCE baseline & Rollout baseline \cite{kool2018attention} \\
        Learning rate & 1e-4 \\
        The Number of Encoder Layer & 3 \\
        Embedding Dimension & 128 \\
        Batch-size & 512 \\
        Epochs & 100\\
        Epoch size & 1,280,000\\
        The Number of Steps & 250$K$ \\
        \hline
    \end{tabular}
    \caption{Hyperparameter Setting for AM for all tasks.}
    \label{tab:my_label}
\end{table}

\textbf{AM.} The AM is a general-purpose DRL-NCO, a transformer-based encoder-decoder model that solves various routing problems such as TSP, CVRP, PCTSP, and OP. We follow the instruction of open source code, same with the PointerNet with the following hyperparameters.

\begin{table}[h]
    \centering
    \begin{tabular}{lc}
        \hline
        REINFORCE baseline & Rollout baseline \cite{kool2018attention} \\
        Learning rate & 1e-4 \\
        The Number of Encoder Layer & 3 \\
        Embedding Dimension & 128 \\
        Attention Head Number & 8 \\
        Feed Forward Dimension & 512 \\
        Batch-size & 512 \\
        Epochs & 100\\
        Epoch size & 1,280,000\\
        The Number of Steps & 250$K$ \\
        \hline
    \end{tabular}
    \caption{Hyperparameter Setting for AM for all tasks.}
    \label{tab:my_label}
\end{table}

\textbf{POMO.} The POMO is a high-performance DRL-NCO for TSP and CVRP, implemented on the top of the AM. We follow the instruction of open source code \footnote{https://github.com/yd-kwon/POMO} with the following hyperparameters. 

\begin{table}[h]
    \centering
    \begin{tabular}{lcc}
        \hline
        &TSP&CVRP\\
        \hline
        REINFORCE baseline & \multicolumn{2}{c}{POMO shared baseline \cite{kwon2020pomo}} \\
        Learning rate & \multicolumn{2}{c}{1e-4}\\
        Weight decay & \multicolumn{2}{c}{1e-6}\\
        The Number of Encoder Layer & \multicolumn{2}{c}{6} \\
        Embedding Dimension & \multicolumn{2}{c}{128} \\
        Attention Head Number & \multicolumn{2}{c}{8} \\
        Feed Forward Dimension & \multicolumn{2}{c}{512} \\
                Batch-size & \multicolumn{2}{c}{64} \\
        Epochs & 2,000 & 8,000 \\
        Epoch size & 100,000 & 10,000\\
        The Number of Steps & 3.125$M$ & 1.25$M$ \\
        \hline
    \end{tabular}
    \caption{Hyperparameter Setting for POMO in TSP and CVRP.}
    \label{tab:my_label}
\end{table}


\clearpage

\section{Implementation Details of Proposed Method}
\label{sec:implement}
\subsection{Training Hyperparameters}
\label{append: hyperparameter}

Sym-NCO is a training scheme that is attached to the top of the existing DRL-NCO model. We set the same hyperparameters with PointerNet, AM, and POMO \cref{append: baseline} except REINFORCE baseline (we set the proposed Sym-NCO baseline introduced in \cref{sec:sym-nco}).

Sym-NCO has additional hyperparameters. First of all, we set identical hyperparameters for PointerNet and AM for all tasks:

\begin{table}[h]
    \centering
    \begin{tabular}{lc}
        \hline
        $\alpha$ & 0.1 \\
        $\beta$ & 0 \\
        $K$ & 1 \\
        $L$ & 10 \\
        \hline
    \end{tabular}
    \caption{Hyperparameter Setting of Sym-NCO for PointerNet and AM.}
    \label{tab:my_label}
\end{table}

Note that the design choice of $\beta = 0$ is to show high applicability of $\psl$, and is because AM with $\ssl$ is just similar to the POMO.  

For POMO, we set $\beta = 1$ to force solution symmetricity on the top of POMO's baseline. Note that we follow POMO's first node restriction only in TSP, which is a reasonable bias as we mentioned in \cref{sec:sym-nco}. The hyperparameter setting is as follows: 

\begin{table}[h]
    \centering
    \begin{tabular}{lcc}
    \hline
    &TSP & CVRP\\
        \hline
        $\alpha$ & 0.1 & 0.2\\
        $\beta$ & 1 & 1\\
        $K$ & 100 & 100\\
        $L$ & 2 & 2\\
        \hline
    \end{tabular}
    \caption{Hyperparameter Setting of Sym-NCO for POMO.}
    \label{tab:my_label}
\end{table}

Note that the design choice of $\beta = 1$ and $K=100$ is based on POMO's baseline setting. We just set $L=2$, because of training efficiency. We suggest to set $L>4$ if training resources and time-budget is sufficient;; it may increase performance further.

\subsection{Integration of $\ssl$ and $\psl$}
\label{append:deisgn_guide}
The $\psl$ is an extension of $\ssl$ where it can both leverage problem symmetricity and solution symmetricity. Therefore, we can simply use $\mathcal{{L}_{\text{RL-Sym}}} = \psl$. However, some specific CO problem such as TSP has cyclic nature, which contains pre-identifiable solution symmetricity, and some method already exploit the cyclic nature. For example, the POMO \cite{kwon2020pomo} which is a powerful NCO model already utilizes pre-identified solution symmetricity in the training process for specific CO tasks. Therefore, we provide a general loss term $\mathcal{{L}_{\text{RL-Sym}}} = \psl + \beta \ssl$ that can be used with POMO or similar methods for specific CO problems (TSP and CVRP). If the problem has pre-identified solution symmetricity (TSP) or has a strong cyclic nature (CVRP), we can set $\beta = 1$ to leverage solution-symmetricity more. If we do not have specific domain knowledge for the target task, then we leave $\beta = 0$, to leverage problem-symmetricity and solution-symmetricity simultaneously using only $\psl$.

\subsection{Multi-start Post-processing}
\label{append:mutl}
To sample multi solutions from one solver $F_{\theta}$ we suggest instance augmentation method following \cite{kwon2020pomo}. As suggested in \cite{kwon2020pomo}, we can generate multiple samples to ablate first node selection of decoding step by $N$. Moreover we can generate $8$ samples to rotate with $0,90,180,270$ degrees with reflection: $4 \times 2 = 8$. To comparison with Sym-NCO and POMO as shwon in \cref{figure:time_perf_analysis} (three markers), we conduct these multi state post processing with sampling width: $1,100,100\times 8$.

We, on the other hand, suggest an extended version of the instance augmentation method of \cite{kwon2020pomo}, using random orthogonal matrix $Q$. By transforming input problem $\boldsymbol{P}$ with $Q^{1},..., Q^{M}$ which are orthogonal matrices, we can sample multiple sample solutions from the $M$ symmetric problems. We used these strategies in PCTSP and OP by setting $M=200$. 

\subsection{Details of Projection Head}

The projection head introduced in \cref{subsec:invariant} is a simple two-layer perception with the ReLU activation function, where input/output/hidden dimensions are equals to encoder's embedding dimension (i.e. 128).

\subsection{Computing Resources and Computing Time}

For training Sym-NCO, we use \textit{NVIDIA} A100 GPU. Because POMO implementation does not support GPU parallelization, we use a single GPU for the POMO + Sym-NCO. It takes approximately two weeks to finish training POMO + Sym-NCO. For training AM + Sym-NCO, we use $4 \times$ GPU, which takes approximately three days to finish training. 

As mentioned in \cref{sec:exp_setting}, we use \textit{NVIDIA} RTX2080Ti single GPU at the test time.

\label{append:multi-start}
\clearpage

\section{Additional Experiments}
\subsection{Hyperparameter Tuning of $\alpha$ in CVRP}
\label{append:hyper-tuinig}

We did not tune hyperparameter much in this work because training resources were limited where Sym-NCO must be verified on several tasks and DRL-NCO architectures. Therefore, we only contain simple hyperparameter ablation for $\alpha$ in CVRP (POMO + SymNCO setting). 

\begin{figure}[h]
         \centering
         \includegraphics[width=0.5\textwidth]{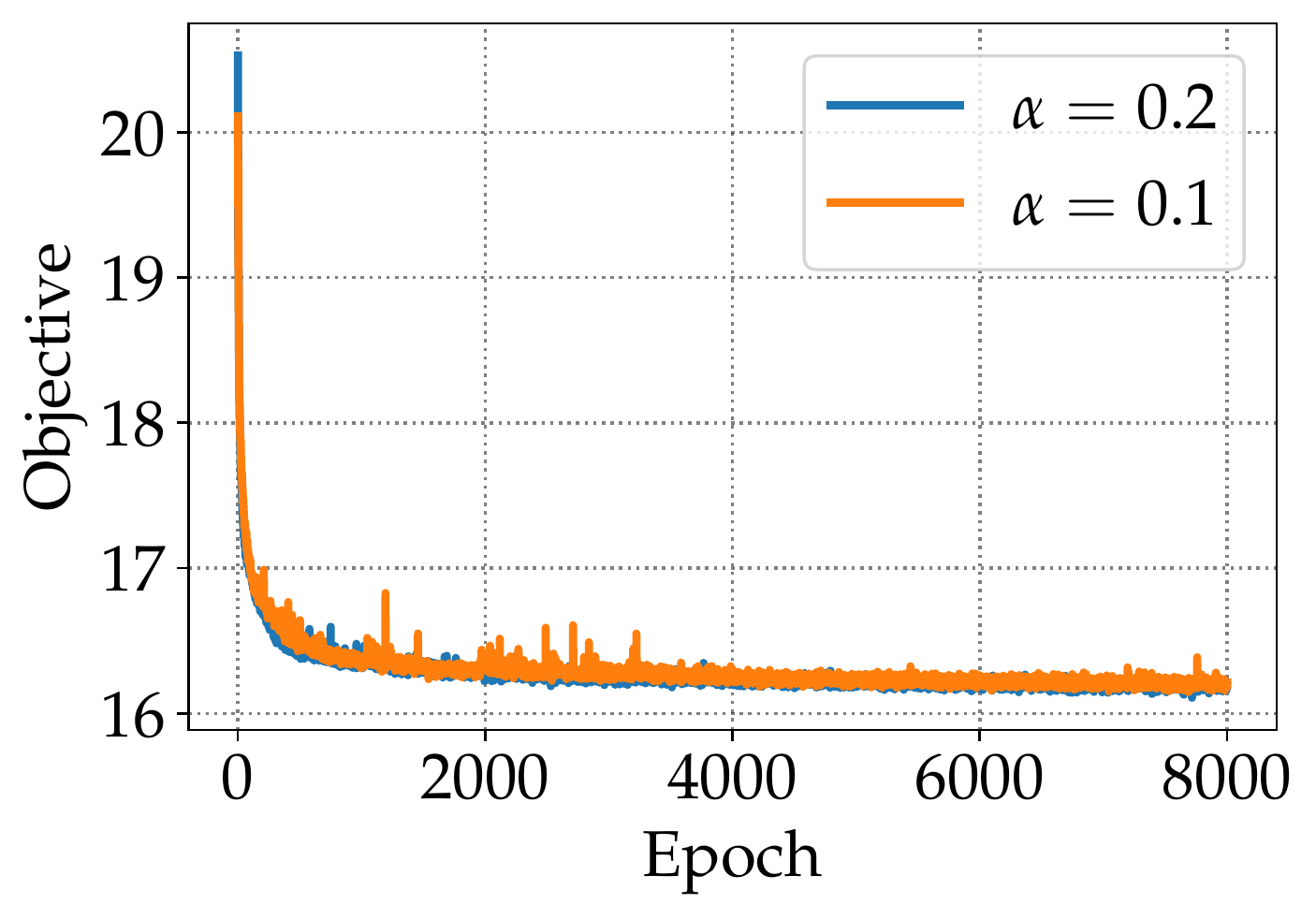}
         \caption{Alblation Study for $\alpha \in \{0.1,0.2\}$}
    \label{figure:hyperparameter}
\end{figure}

This validation results shows $\alpha = 0.2$ give slightly better performances than $\alpha = 0.1$, but tuning of $\alpha$ semms to be not sensitive.  

\clearpage
\subsection{Performance Evaluation on TSPLIB }

This section gives Sym-NCO performance evaluation in the TSPLIB ($N<250)$. Sym-NCO and the POMO is pre-trained model in $N=100$ that is evaluated in \cref{table:tsp_cvrp_100}. In this experiment, we conduct multi-start sampling with sample width $M=N \times 20$ where the $N$ indicates multi initial city sampling of problem size (ex., the ``eil51" has $N=51$). The $20$ indicates multi-sampling using random orthogonal matrix as we introduced in \cref{append:multi-start}. As shown in the below table, our Sym-NCO outperforms POMO, having a $1.62\%$ optimal gap, which is extremely high performance in real-world TSPLIB evaluation compared with other NCO evaluations \cite{kim2021learning}.

\begin{table}[h]
\begin{center}
\fontsize{9}{9}\selectfont
     \caption {Performance comparison in real-world instances in TSPLIB.\\ }
     \label{tab:temps}
     
\begin{tabular}{llllll}

\specialrule{1pt}{0pt}{4pt}
\multicolumn{1}{l}{\begin{tabular}{c}\multirow{2}{*}{Instance}\end{tabular}}&\multicolumn{1}{c}{\begin{tabular}{cc}\multirow{2}{*}{Opt.}\end{tabular}}&\multicolumn{2}{c}{POMO \cite{kwon2020pomo}}&\multicolumn{2}{c}{Sym-NCO (ours)}\\\cmidrule[0.5pt](lr{0.2em}){3-4} \cmidrule[0.5pt](lr{0.2em}){5-6} 
\multicolumn{2}{c}{}&\multicolumn{1}{c}{Cost}&\multicolumn{1}{c}{Gap}&\multicolumn{1}{c}{Cost}&\multicolumn{1}{c}{ Gap}\\
\specialrule{0.7pt}{4pt}{4pt}
eil51  & {426}          & {429}     & {0.82$\%$}              & {432}     & {1.39$\%$} \\
berlin52  & {7,542}     & {7,545}   & {0.04$\%$}             & {7,544}    & {0.03$\%$}    \\
st70  & {675}           & {677}     & {0.31$\%$}              & {677}     & {0.31$\%$}    \\
pr76  & {108,159}       & {108,681} & {0.48$\%$}              & {108,388} & {0.21$\%$}    \\
eil76  & {538}          & {544}     & {1.18$\%$}              & {544}     & {1.18$\%$}  \\
rat99  & {1,211}        & {1,270}   & {4.90$\%$}              & {1,261}   & {4.17$\%$}   \\
rd100  & {7,910}        & {7,912}   & {0.03$\%$}              & {7,911}   & {0.02$\%$}    \\
KroA100  & {21,282}     & {21,486}  & {0.96$\%$}              & {21,397}  & {0.54$\%$}    \\
KroB100  & {22,141}     & {22,285}  & {0.65$\%$}              & {22,378}  & {1.07$\%$}   \\
KroC100  & {20,749}     & {20,755}  & {0.03$\%$}              & {20,930}  & {0.87$\%$}    \\
KroD100  & {21,294}     & {21,488}  & {0.91$\%$}              & {21,696}  & {1.89$\%$}    \\
KroE100  & {22,068}     & {22,196}  & {0.58$\%$}              & {22,313}  & {1.11$\%$}    \\
eil101  & {629}         & {641}     & {1.84$\%$}              & 641       & {1.84$\%$}   \\
lin105  & {14,379}      & {14,690}  & {2.16$\%$}              & {14,358}  & {0.54$\%$}   \\
pr124 & {59,030}        & {59,353}  & {0.55$\%$}              & 59,202    & {0.29$\%$}   \\
bier127  & {118,282}    & {125,331} & {5.96$\%$}              & 122,664   & {3.70$\%$}    \\
ch130  & {6,110}        & {6,112}   & {0.03$\%$}              & {6,118}   & {0.14$\%$}    \\
pr136  & {96,772}       & {97,481}  & {0.73$\%$}              & {97,579}  & {0.83$\%$}   \\
pr144  & {58,537}       & {59,197}  & {1.13$\%$}              & {58,930}  & {0.67$\%$}    \\
kroA150  & {26,524}     & {26,833}  & {1.16$\%$}              & {26,865}  & {1.28$\%$}    \\
kroB150  & {26,130}     & {26,596}  & {1.78$\%$}              & {26,648}  & {1.98$\%$}    \\
pr152  & {73,682}       & {74,372}  & {0.94$\%$}              & 75,292    & {2.18$\%$}   \\
u159  & {42,080}        & {42.567}  & {1.16$\%$}              & 42,602    & {1.24$\%$}    \\
rat195  & {2,323}       & 2,546     & {9.58$\%$}              & {2,502}   & {7.70$\%$}   \\
kroA200  & {29,368}     & {29,937}  & {1.94$\%$}             & {29,816}  & {1.53$\%$}  \\
ts225  & {126,643}      & {131,811} & {4.08$\%$}              & 127,742   & {0.87$\%$}    \\
tsp225  & {3,919}       & {4,149}   & {5.87$\%$}             & 4,126     & {5.27$\%$}    \\
pr226  & {80,369}       & {82,428}  & {2.56$\%$}             & {82,337}  & {2.45$\%$}    \\
\specialrule{0.7pt}{4pt}{4pt}
\text{Avg Gap}&{0.00$\%$}&\multicolumn{2}{c}{1.87$\%$}&\multicolumn{2}{c}{\textbf{1.62}$\%$}\\ 
\specialrule{1pt}{4pt}{0pt}
\end{tabular}     
\end{center}     
     
\end{table}

\label{append:TSPLIB}

\label{append:TSPLIB}

\clearpage

\subsection{Performance Evaluation of Transferability to Large Scale Problems}
\label{large_scale}
This section verifies that the pre-trained model using the Sym-NCO has powerful transferability on large-scale problems. We use the efficient-active-search (EAS) \cite{hottung2021efficient} as a transfer learning algorithm for large-scale TSP \footnote{All the hyperparameters are the same with https://github.com/ahottung/EAS}. In transfer learning, the number of iterations is an important factor for adaptation. We set the iteration $K=200$ as default; we provide an ablation study for few-shot learning $K \in \{1,2,5,10\}$ to show Sym-NCO's few-shot adaptation capability. Note that the pre-trained model is trained on CVRP ($N=100$).

As shown in \cref{tab:large_scale}, our method outperforms the POMO \cite{kwon2020pomo} in large-scale CVRP, having only a small performance gap with LKH3. Furthermore, our method increase few shot adaptation capabilities for large-scale tasks, where our model achieved better performances than POMO with 5 $\times$ reduced training shot $K$. To sum up, our Sym-NCO can be positioned with an effective pretraining scheme that approximately imposes symmetricity and is further transferred to larger-scale tasks. 

\begin{table}[h]
\begin{center}
\fontsize{9}{9}\selectfont
     \caption {Performance comparison in large scale CVRP. The performance is evaluated on ten random generated CVRP data. \\ }
     \label{tab:large_scale}
\begin{tabular}{lcccc}

\specialrule{1pt}{0pt}{4pt}
&\multicolumn{2}{c}{CVRP ($N=500$)}&\multicolumn{2}{c}{CVRP ($N=1,000$)}\\\cmidrule[0.5pt](lr{0.2em}){2-3}\cmidrule[0.5pt](lr{0.2em}){4-5} 
\multicolumn{1}{c}{}&\multicolumn{1}{c}{Cost}&\multicolumn{1}{c}{Gap}&\multicolumn{1}{c}{Cost}&\multicolumn{1}{c}{Gap}\\
\specialrule{0.7pt}{4pt}{4pt}
    LKH3 \cite{lkh2017} & 60.37 & 0.00\% & 115.74 & 0.00\%\\
    \specialrule{0.7pt}{4pt}{4pt}
    POMO \cite{kwon2020pomo} + EAS \cite{hottung2021efficient} & 63.30 & 4.85\% & 126.56& 9.34\%\\
    Ours + EAS \cite{hottung2021efficient} & \textbf{62.41} & \textbf{3.37}\%& \textbf{121.85}&\textbf{5.92}\%\\

\specialrule{1pt}{4pt}{0pt}
\end{tabular}     
\end{center}     
     
\end{table}

\begin{table}[h]
\begin{center}
\fontsize{9}{9}\selectfont
     \caption {Performance evaluation of few shot adaptation to large scale CVRP. \\ }
     \label{tab:few_shot}
\begin{tabular}{lcccccccc}
\specialrule{1pt}{0pt}{4pt}
&\multicolumn{4}{c}{CVRP ($N=500$)}&\multicolumn{4}{c}{CVRP ($N=1,000$)}\\\cmidrule[0.5pt](lr{0.2em}){2-5}\cmidrule[0.5pt](lr{0.2em}){6-9} 
&$K=1$&$K=2$&$K=5$&$K=10$&$K=1$&$K=2$&$K=5$&$K=10$\\
\specialrule{0.7pt}{4pt}{4pt}
POMO \cite{kwon2020pomo} + EAS \cite{hottung2021efficient} & 136.91&116.77&77.57&69.90&366.61&311.41&189.26&162.64\\
Sym-NCO + EAS \cite{hottung2021efficient}& \textbf{75.85}&\textbf{69.71}&\textbf{67.26}&\textbf{66.33}&\textbf{192.12}&\textbf{163.92}&\textbf{139.66}&\textbf{134.61}\\
\specialrule{1pt}{4pt}{0pt}
\end{tabular}     
\end{center}     
     
\end{table}

\clearpage

\subsection{Comparison with Deep Improvement Heuristic Methods}


In this section, we provide performance comparison with state-of-the-art deep improvement heuristics. 
As shown in \cref{tab:improvement}, our method outperformed state-of-the-art deep improvement heuristics with the fastest speed. Note that constructive heuristics (which include our method) and improvement heuristics are complementary and can support each other.  

\begin{table}[h]
\begin{center}
\fontsize{9}{9}\selectfont
     \caption {Performance comparison with deep improvement heuristics. The $I$ indicates the number of iterations, and the $s$ indicates the number of samples per instance. \\ }
     \label{tab:improvement}
\begin{tabular}{lcccc}

\specialrule{1pt}{0pt}{4pt}
&\multicolumn{2}{c}{TSP ($N=100$)}&\multicolumn{2}{c}{TSP ($N=100$)}\\\cmidrule[0.5pt](lr{0.2em}){2-3}\cmidrule[0.5pt](lr{0.2em}){4-5} 
\multicolumn{1}{c}{}&\multicolumn{1}{c}{Cost}&\multicolumn{1}{c}{Gap}&\multicolumn{1}{c}{Cost}&\multicolumn{1}{c}{Gap}\\
\specialrule{0.7pt}{4pt}{4pt}
    Wu et al. ($I=5K$) \cite{wu2020learning} (I=5K)& 1.42\% & 2h & 2.47\% & 5h\\
    DACT ($I=1K$) \cite{ma2021learning} & 1.62\% & 48s & 3.18\%& 2m\\
    DACT ($I=5K$) \cite{ma2021learning} & 0.61\% & 4m &1.55\%&8m \\
    Ours ($s.100$)  & \textbf{0.39}\% &12s& \textbf{1.46}\%&16s \\
    Ours ($s.800$)  & \textbf{0.14}\% &1m& \textbf{0.90}\%&2m\\
\specialrule{1pt}{4pt}{0pt}
\end{tabular}     
\end{center}     
     
\end{table}

We remark that the speed evaluation of Wu et al. \cite{wu2020learning} and DACT \cite{ma2021learning} is referred to by \cite{ma2021learning} where the speed is evaluated with NVIDIA TITAN RTX. The speed of our method is evaluated with NVIDIA RTX 2080Ti. 

\clearpage

\subsection{Comparison with Symmetric NCO models}
\label{sym_literature}
\textbf{Previous Symmetricitcy Considered NCO methods vs. Sym-NCO.} Several studies exploited the symmetric nature of CO. Ouyang et al. \cite{ouyang2021generalization} have a similar purpose to Sym-NCO in that both are DRL-based constructive heuristics, but they give rule-based input transformation (relative position from first visited city) to satisfy equivariance. However, our method learns to impose symmetricity approximately into the neural network with regularization loss term. We believe our approach is a more general approach to tackling symmetricity (see \cref{tab:sym_func}) because not every task can be represented as a relative position with the first visited city.

The Hudson et al. \cite{hudson2021graph} is the extended work of Joshi et al. \cite{joshi2020learning} where graph neural network (GNN) makes a sparse graph from a fully connected input graph, and the search method figures out the feasible solution from the sparse graph. This method is based on the supervised learning scheme that requires expert labels. Moreover, this method does not guarantee to generate feasible solutions in hard-constraint CO tasks because the pruning process of the GNN may eliminate feasible trajectory (In TSP, it may work, but in other tasks, this method must address feasibility issues). Regardless of this limitation, we view the line graph transformation suggested by Hudson et al. \cite{hudson2021graph} as novel and helpful in terms of symmetricity.   

Ma et al. \cite{ma2021learning} proposed a DRL-based improvement heuristic, exploiting the cyclic nature of TSP and CVRP. The purpose of Ma et al.\cite{ma2021learning}, and our Sym-NCO is different: the objective of Sym-NCO is approximately imposing symmetricity nature, but the objective of Ma et al. \cite{ma2021learning} is to improve the iteration process of improvement heuristic with fined designed positional encoding for TSP and CVRP. Note that Sym-NCO (constructive method) and Ma et al. \cite{ma2021learning} (Improvement method) are complementary and can support each other. For example, pretrained constructive model can generate an initial high-quality solution, whereas an improvement method can iteratively improve solution quality. 

\textbf{Experimental Comparison.}

Our Sym-NCO outperforms all the relevant related baselines as shown in \cref{tab:sym_perf}. Furthermore, \cref{tab:sym_func} shows our method covers the widest arrange of CO tasks, where it does not needs labeled data.  

\begin{table}[h]
\begin{center}
\fontsize{9}{9}\selectfont
     \caption {Performance comparison with symmetric NCO methods\\ }
     \label{tab:sym_perf}
\begin{tabular}{lccc}
\specialrule{1pt}{0pt}{4pt}
&Optimal Gap&Time&GPU resources\\
\specialrule{0.7pt}{4pt}{4pt}
Ouyang et al. \cite{ouyang2021generalization}& 2.61\% &1.3m & GTX1080Ti\\
Hudson et al. \cite{hudson2021graph}& 0.698\% &28h & Tesla P100\\
Ma et al. \cite{ma2021learning}& 1.62\% &48s & Titan RTX\\
Ours &\textbf{0.39}\%&\textbf{12s}&RTX 2080Ti\\
\specialrule{1pt}{4pt}{0pt}
\end{tabular}     
\end{center}     
\end{table}

\begin{table}[h]
\begin{center}
\fontsize{9}{9}\selectfont
     \caption {Performance comparison with symmetric NCO methods\\ }
     \label{tab:sym_func}
\begin{tabular}{lcc}
\specialrule{1pt}{0pt}{4pt}
&Learning Methods&Verified Tasks\\
\specialrule{0.7pt}{4pt}{4pt}
Ouyang et al. \cite{ouyang2021generalization}& Reinforcement Learning & TSP\\
Hudson et al. \cite{hudson2021graph} & Supervised Learning & TSP\\
Ma et al. \cite{ma2021learning}& Reinforcement Learning & TSP, CVRP\\
Ours &Reinforcement Learning& TSP, CVRP, PCTSP, OP\\
\specialrule{1pt}{4pt}{0pt}
\end{tabular}     
\end{center}     
\end{table}

\end{document}